\documentclass[11pt]{article}

\usepackage[preprint]{acl}

\usepackage{times}
\usepackage{latexsym}
\usepackage{amsmath}
\usepackage{booktabs}
\usepackage{multirow}
\usepackage[table]{xcolor}
\usepackage{enumitem}

\usepackage[T1]{fontenc}

\usepackage[utf8]{inputenc}

\usepackage{microtype}

\usepackage{inconsolata}

\usepackage{graphicx}

\title{The Point, the Vision and the Text: Does Point Cloud Boost Spatial Reasoning of Large Language Models? A Bias-Controlled Study}

\author{
\textbf{Weichen Zhang}$^{1}$,
\textbf{Ruiying Peng}$^{1}$,
\textbf{Xin Zeng}$^{1}$,
\textbf{Jianjie Fang}$^{1}$,
\textbf{Ziyou Wang}$^{1}$,
\textbf{Kaiyuan Li}$^{1}$,
\\
\textbf{Heng Dong}$^{2}$,
\textbf{Wei Li}$^{2}$,
\textbf{Chen Gao}$^{1,\ddagger}$,
\textbf{Xin Wang}$^{1}$,
\textbf{Xinlei Chen}$^{1,\ddagger}$,
\textbf{Yong Li}$^{1}$
\\[2mm]
$^{1}$Tsinghua University
\quad
$^{2}$ByteDance Seed
\quad
$^{\ddagger}$Corresponding Author
\\
\texttt{chgao96@gmail.com},
\texttt{chen.xinlei@sz.tsinghua.edu.cn}
}

\begin{document}
\maketitle

\begin{abstract}
3D Large Language Models (LLMs) leveraging spatial information in point clouds for 3D spatial reasoning attract great attention. Despite some promising results, the advantages of point clouds over other modalities remain unclear. Moreover, existing 3D benchmarks are insufficient for fairly evaluating the ability of multimodal LLMs to comprehend spatial concepts. To address these challenges, we introduce ScanReQA, a 3D spatial reasoning benchmark encompassing text, vision, and point cloud modalities. We then evaluate the performance of text, 2D, and 3D LLMs on the benchmark to compare the effectiveness of different modalities in understanding spatial concepts. Furthermore, we analyze the reasoning mechanisms behind 3D LLMs using point clouds. Our findings reveal that: 1) binary spatial reasoning remains challenging for current 3D LLMs, 2) MLLMs based on point cloud and visual modalities demonstrate stronger spatial reasoning capabilities than LLMs, and 3) 3D LLMs exhibit the attention sink phenomenon similar to that in 2D LLMs, impairing spatial reasoning.
We think these conclusions can help the next step of 3D LLMs and also offer insights for foundation models in other modalities.
We release datasets and codes in the project page: \url{https://github.com/EmbodiedCity/ScanReQA.code}.



\end{abstract}    
\begin{figure*}[t!]
  \centering
  \includegraphics[width=1.0\linewidth]{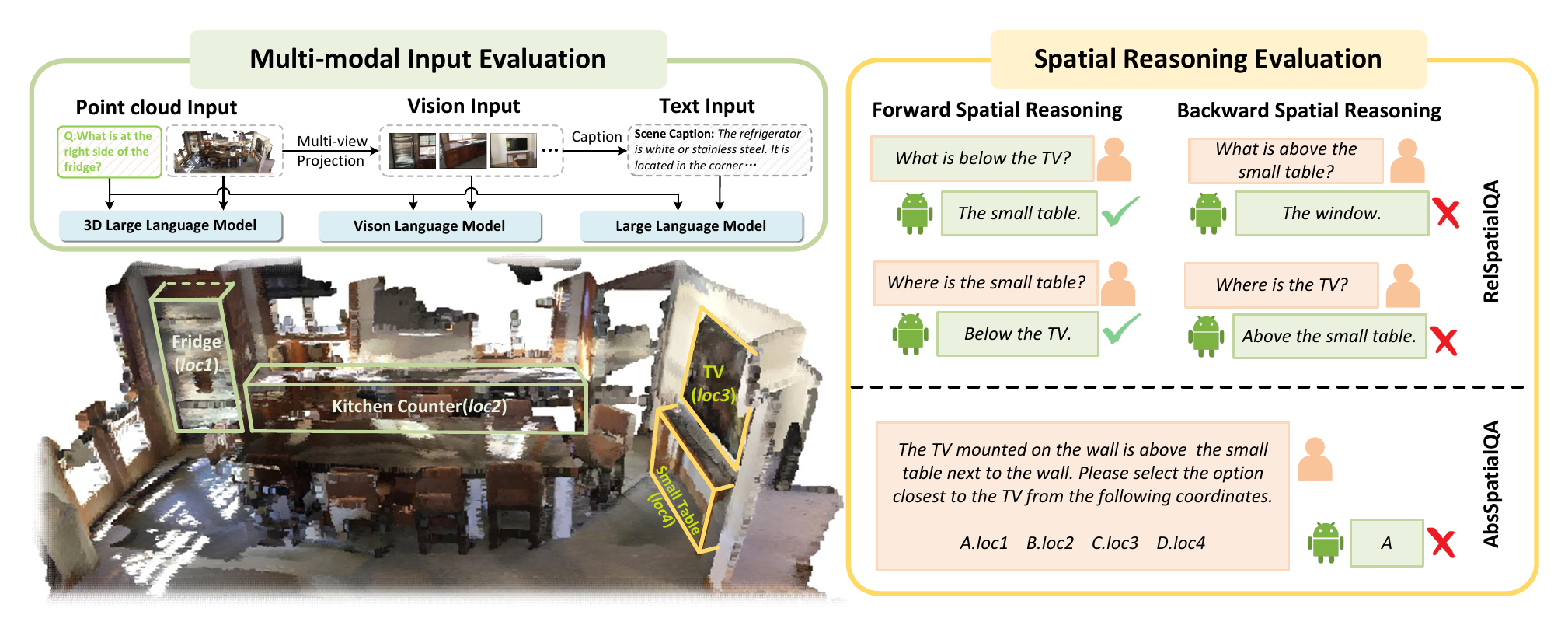}
   \vspace{-1cm}
      \caption{\textbf{The overview of our evaluation framework on 3D LLMs.} In the multimodal input evaluation, we generate point cloud inputs along with their visual and textual counterparts, which are fed into the corresponding models. For spatial reasoning evaluation, 3D LLMs are tasked with both forward and backward spatial reasoning, as well as predicting the location of the referenced object. \textit{loc}\# is the 3D coordinate of the corresponding object.}
  \vspace{-0.4cm}
  \label{fig:cover}
\end{figure*}

\section{Introduction}
\label{sec:intro}

Equipping models with spatial reasoning is essential for bridging the gap between artificial and human intelligence and is a critical step toward general intelligence. While LLMs~\cite{touvron2023llama, achiam2023gpt, brown2020language, bai2023qwen} and vision-language models (VLMs)~\cite{zhu2024llava, bai2023qwen, team2023gemini} have demonstrated strong performance in tasks such as commonsense reasoning~\cite{davis2023benchmarks}, math problem solving~\cite{liu2025matheval}, and scientific QA~\cite{lu2022learn}, they remain limited in spatial reasoning due to the absence of 3D perception. To address this, recent efforts incorporate 3D spatial information from point clouds into LLMs~\cite{zhu20233d, man2024lexicon3d}, enabling progress in tasks such as 3D captioning, spatial QA, and embodied navigation.

Intuitively, 3D LLMs equipped with spatial cues from point clouds should outperform text-only LLMs or VLMs in spatial reasoning. However, existing results tell a different story. Current 3D LLMs achieve only around 50\% accuracy on 3D QA benchmarks~\cite{azuma2022scanqa, ma2022sqa3d}, falling short of SOTA results in visual~\cite{liu2024mmbench} and text-based QA~\cite{hendrycks2020measuring}. Moreover, they often fail to generalize spatial relationships across scene variations~\cite{man2024situational}, raising concerns about \textit{whether point cloud inputs genuinely enhance the spatial reasoning capabilities of 3D LLMs}.

To answer this question, there are two challenges to overcome: 1) how to evaluate the spatial reasoning capacities of 3D LLMs? Although current 3D benchmarks include various spatial reasoning tasks~\cite{azuma2022scanqa,ma2022sqa3d,zhi2025lscenellm}, the spatial relationships in these benchmarks often exhibit a bias towards object semantics. Some spatial relationships consistently co-occur with specific objects~\cite{peng2025understanding}, causing models to overfit to these fixed spatial relationships when encountering similar objects.
2) How can consistent scene information be represented across text, vision, and point cloud modalities to enable a fair comparison of spatial reasoning abilities among text, 2D, and 3D LLMs? It is essential to ensure that all three modalities convey the necessary spatial reasoning cues while not introducing information not grounded in the original scene.

To address the first challenge, we construct ScanReQA, an unbiased 3D QA benchmark based on ScanQA~\cite{azuma2022scanqa}, the open-sourced spatial QA benchmark, as shown in Figure~\ref{fig:cover}. ScanReQA focuses on binary spatial relationships in ScanQA that can be represented by reversible spatial concepts (e.g., \textit{``left"} and \textit{``right"}) and a pair of objects. We design both forward and reverse spatial reasoning questions for each object pair, such as \textit{``what is A to B"} and \textit{``what is B to A"}. 
This design ensures that forward and reversed spatial relationships occur with equal frequency across object pairs, mitigating biases in both object and relationship distributions.
Notably, ScanReQA is \textbf{NOT} an independent benchmark from ScanQA; rather, it complements ScanQA by specifically addressing the bias caused by the co-occurrence of spatial relationships with particular objects.

To address the second challenge, we design an evaluation framework as shown in Figure~\ref{fig:cover}. All scene representations across different modalities are derived from original scene videos, with point cloud and textual representations generated through 3D projection and image captioning, respectively. Subsequently, we solely input each modality’s scene representation along with the corresponding questions into the respective LLMs to compare the performance differences. 



In this work, we introduce an unbiased 3D QA benchmark for binary spatial relationship reasoning and conduct a systematic evaluation of state-of-the-art multimodal LLMs. We find that 3D LLMs with point cloud input tend to outperform zero-shot LLMs and VLMs by at least 3\%, while underperforming fine-tuned VLMs by approximately 8\%. Moreover, their accuracy on bidirectional spatial reasoning tasks remains below 10\%.

To better understand these limitations, we perform an in-depth analysis of how 3D LLMs process point cloud inputs by examining attention distributions and semantic token flow across model layers. Through attention visualization and logit-lens analysis, we identify attention sink and object semantic misalignment phenomena in current 3D LLMs, providing insights into the challenges and potential directions for improving their spatial reasoning capabilities.

The main contributions are summarized below: 
\begin{itemize}[leftmargin=*,partopsep=0pt,topsep=0pt]
\setlength{\itemsep}{0pt}
\setlength{\parsep}{0pt}
\setlength{\parskip}{0pt}
    \item We present the first systematic evaluation of how multimodal inputs affect 3D spatial reasoning, providing a comprehensive assessment of 3D LLMs' reasoning capabilities.
    
    \item We introduce ScanReQA, the first unbiased 3D QA benchmark designed to evaluate existing 3D LLMs on binary spatial relationship reasoning.

    \item We conduct an in-depth analysis of attention patterns and token-level semantic flow in 3D LLMs, revealing two critical factors that affect spatial reasoning: attention sink and object semantic misalignment.
\end{itemize}

\begin{figure*}[t!]
  \centering
  \includegraphics[width=0.98 \linewidth]{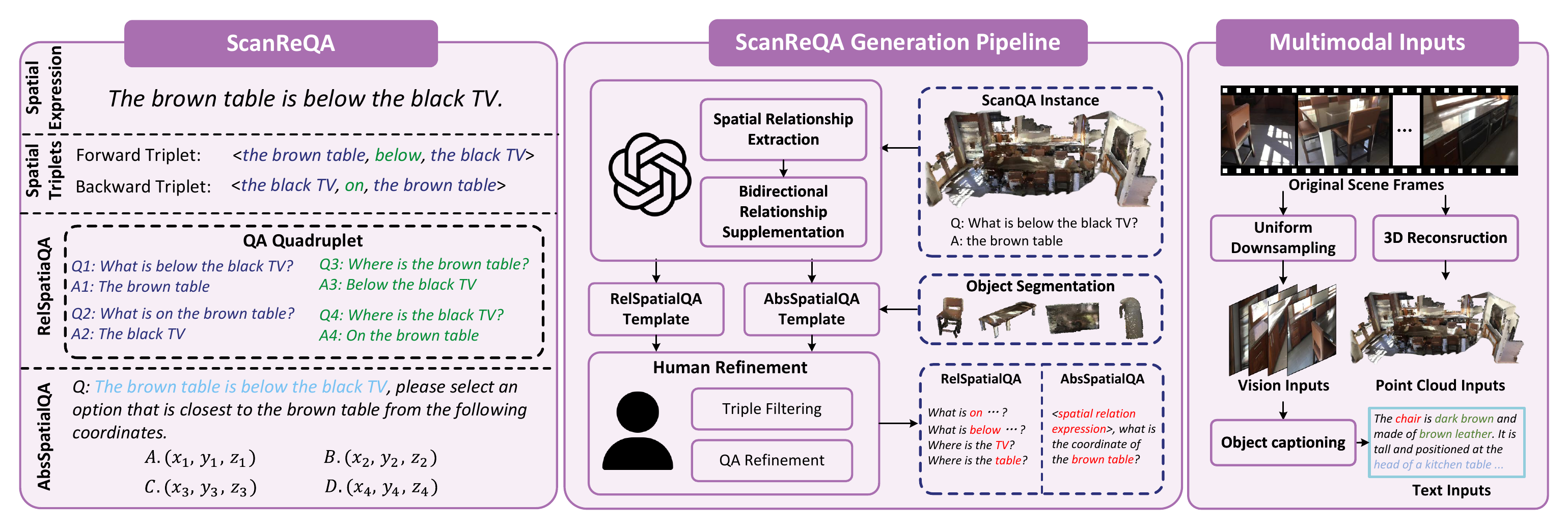}
   \vspace{-0.3cm}
  \caption{\textbf{Overview of ScanReQA and its generation pipeline.} \textit{Left}: ScanReQA is derived from spatial expressions in the ScanQA dataset. Binary spatial relationships are extracted and represented as forward and backward triplets, which are used to construct QA quadruplets in RelSpatialQA and coordinate-based options in AbsSpatialQA. Each QA quadruplet involves questions reasoning about both \textcolor[RGB]{101,102,157}{\textit{what-is}} and their \textcolor[RGB]{4,141,54}{\textit{where-is}} QAs. \textit{Middle}: LLMs extract spatial expressions and generate triplets, which are filled into QA templates and later refined by human annotators. \textit{Right}: Original scene frames are uniformly downsampled and reconstructed into 3D point clouds to produce 2D/3D inputs, while visual inputs are used for object captioning to generate textual descriptions.}
  \vspace{-0.4cm}
  \label{fig:qa_gen}
\end{figure*}
\section{Related Work}
\subsection{Benchmark for Spatial Reasoning}

Spatial reasoning is commonly evaluated via spatial QA accuracy, with existing benchmarks broadly categorized into situated and non-situated QA. 

ScanQA~\cite{azuma2022scanqa}, built on ScanNet~\cite{dai2017scannet}, is an early and representative non-situated benchmark. Subsequent non-situated datasets~\cite{yang20253d,lyu2024mmscan,huang2023embodied} such as Beacon3D~\cite{huang2025unveiling} emphasize densely grounded point cloud–language QA to decouple object grounding from reasoning. Another line of work~\cite{man2024situational,chen2024ll3da,zhu20233d} focuses on situated QA~\cite{ma2022sqa3d}, where providing the observer’s viewpoint reduces ambiguity. MSQA~\cite{linghu2024multi} further replaces textual object descriptions with visual inputs to mitigate relational ambiguity. Real-3DQA~\cite{ma20263d} rotates the observer's orientation to evaluate the robustness of 3D LLMs' capabilities under viewpoint changes.
ScanReQA is a non-situated QA benchmark. It differs from prior benchmarks by emphasizing bidirectional reasoning over binary spatial relations, evaluating whether models that correctly answer forward spatial QAs can also infer their inverse counterparts when both relations are present in the model vocabulary.




\subsection{MLLM on spatial reasoning}



Typical 3D spatial reasoning tasks include captioning, grounding, question answering, dialogue, planning, navigation, and manipulation. Recent multimodal LLMs have shown strong potential for complex spatial reasoning and human–agent interaction. However, while models such as LLaVA~\cite{liu2024visual}, Flamingo~\cite{alayrac2022flamingo}, BLIP-2~\cite{li2023blip}, and PaLM-E~\cite{driess2023palm} perform well in 2D vision-language tasks, they remain limited in 3D understanding and reasoning.

To bridge this gap, recent studies extend LLMs or 2D VLMs to 3D domains by integrating point cloud features with textual inputs, leveraging LLMs’ semantic reasoning ability to generate context-aware outputs. Moreover, the planning, decision-making, and tool-use capabilities of MLLMs make them promising backbones for embodied 3D agents that can navigate~\cite{huang2023embodied,rana2023sayplan,zheng2024towards}, interact with objects~\cite{mirjalili2023lan}, and use tools~\cite{hong2024multiply} while performing complex spatial reasoning.




\section{ScanReQA Design and Construction}
In this part, we first introduce the insights of benchmark design and then clarify the detailed construction pipeline. 
\subsection{Benchmark Design Principles}

Due to factors like dataset distribution differences and biases affecting model performance, it is crucial to eliminate these confounders to ensure that the model's performance accurately reflects its spatial reasoning abilities. Thus, the design of ScanReQA follows three key principles.


\textbf{Data distribution consistency} \textit{This principle eliminates the effect of dataset distribution differences on model performance.}
If the model understands a spatial concept rather than fitting to a specific QA pattern, it should be able to identify the same spatial relationship across different datasets. However, since model performance is affected by dataset distributions, we align ScanReQA with ScanQA in terms of scene selection, spatial relationship types, and question formats. This ensures that any performance differences between the two benchmarks are primarily due to the model's understanding of spatial relationships, rather than differences in dataset distribution. Specifically, ScanReQA aligns with ScanQA by 1) selecting the same 3D scenes from ScanNet~\cite{dai2017scannet}, 2) containing the same spatial relationships, and 3) maintaining consistent question styles and formats.

\textbf{QA Unbiasedness} \textit{This eliminates the model's bias towards specific spatial relationships and objects.}
In current benchmarks such as ScanQA, certain object categories appear with significantly higher frequency than others. A similar imbalance is observed in the distribution of spatial relations. As a result, models trained on ScanQA tend to develop biases toward specific object types and spatial relation patterns. To address this issue, ScanReQA builds upon ScanQA by explicitly constructing QA pairs that cover a diverse range of object categories and spatial relations. This design ensures a more balanced distribution across different types of objects and spatial relations in the dataset, thereby promoting fairness and reducing evaluation bias.


\textbf{Multimodal information consistency} \textit{This eliminates the effect of multimodal information inconsistency on model performance.}
To evaluate and compare the spatial reasoning performance of large models across different modalities, ScanReQA represents scene information using text, vision, and point clouds. To ensure a fair comparison, all modality data is derived from the original ScanNet scene videos. Specifically, point cloud data is obtained through 3D projection, images are generated from different viewpoints via video downsampling, and scene text descriptions are created through image captioning. This approach ensures consistency across modalities, thereby guaranteeing a fair evaluation.

\subsection{Benchmark Generation Pipeline}
Our proposed benchmark, ScanReQA, focuses on binary spatial relationships, which reflect the fundamental spatial reasoning capabilities of models. The underlying insight is that if a model understands binary spatial relationships, given "A is left to B," it should be able to reason that "B is right to A." Moreover, understanding binary spatial relationships is a necessary capability for handling more complex spatial relationships.

Generally, a binary spatial relationship can be formalized as a triplet $\left \langle t, r, a\right \rangle$~\cite{dai2017scannet, zhang2023multi3drefer}, where $t$, $r$, and $a$ denote the target object, the spatial relationship phase, and the anchor object, respectively. If the relationship is reversible, it also has an equivalent representation $\left\langle a, r^{-1}, t\right\rangle$. For simplicity, we refer to $\left \langle t, r, a\right \rangle$ as the forward triplet and $\left \langle a, r^{-1}, t\right \rangle$ as the backward triplet. We define the spatial reasoning capacity of 3D LLMs in two aspects: 1) the ability to infer the third element of a forward/backward triplet given any two elements, which is formulated as:
\begin{equation}
\setlength\abovedisplayskip{3pt}
\setlength\belowdisplayskip{3pt}
\label{eq:spatial-reasoning}
\begin{aligned}
    &LLM(a, r; \theta) \Rightarrow t, \quad
    LLM(t, r^{-1}; \theta) \Rightarrow a,\\
    &LLM(a, t; \theta) \Rightarrow r,  \quad
    LLM(t, a; \theta) \Rightarrow r^{-1},
\end{aligned}
\end{equation}
where $\theta$ is the model weights.
2) The ability to infer the target object's coordinates given a spatial triplet, which is formulated as: 
\begin{equation}
\setlength\abovedisplayskip{3pt}
\setlength\belowdisplayskip{3pt}
\label{eq:spatial-abs}
\begin{aligned}
    &LLM(a, r, t; \theta) \Rightarrow C_t, \\
    &LLM(t, r^{-1}, a; \theta) \Rightarrow C_a,
\end{aligned}
\end{equation}
where $C_t$ and $C_a$ denote the numerical coordinates of the target and anchor objects, respectively.
To assess these two capabilities, ScanReQA introduces two distinct QA datasets: RelSpatialQA and AbsSpatialQA, which contain QAs about direction and coordinate spatial reasoning. The complete generation pipeline is depicted in Figure~\ref{fig:qa_gen}.

\textbf{Binary Spatial Relationship Extraction} Following the first benchmark design principle outlined in the previous section, ScanReQA is constructed based on the binary spatial relationships extracted from ScanQA. Specifically, we prompt the LLM to: 1) convert the QA pair into a spatial relationship expression, as illustrated in Figure~\ref{fig:qa_gen}, 2) verify the reversibility of the spatial relationship, and 3) represent the relationship as a triplet, referred to as a spatial triplet.

\textbf{Bidirectional Relationship Supplementation} Following the second principle, we supplement each extracted relationship with its reversible counterpart. For example, if the original triplet is $\left\langle A, left, B\right\rangle$, the reversed triplet would be $\left\langle B, right, A\right\rangle$. We then query both the relationship of $A$ to $B$ and $B$ to $A$, as well as what is to the left of $B$ and what is to the right of $A$. In this case, each binary spatial relationship corresponds to four QAs. This ensures that the frequency of binary relationships and objects in the answers remains balanced.

\textbf{Template-based QA generation} We follow the first principle to design four QA templates for RelSpatialQA and one template for AbsSpatialQA based on the ScanQA style. The QA templates are shown in Figure~\ref{fig:qa_gen}. RelSpatialQA templates consists of two forward and two backward spatial reasoning QAs, where each reasoning direction includes one \textit{what-is} QA and one \textit{where-is} QA.

\textbf{Multi-modal input generation} 
Following the last principle, we generate scene representations in text, vision, and point cloud modalities, all derived from ScanNet scene videos. For video-to-point cloud conversion, we directly use the ScanNet point clouds reconstructed from 3D scenes. For video-to-vision conversion, we uniformly downsample each video to 64 frames to match the input format of video LLMs~\cite{zheng2025video, chen2024internvl}. For video-to-text conversion, we follow SceneVerse~\cite{jia2025sceneverse} by projecting object point clouds onto 2D planes to identify their presence in RGB frames, and then employ BLIP2~\cite{li2023blip} to generate object captions. To avoid potential information leakage, we follow MSQA~\cite{linghu2024multi} and represent each scene as a collection of objects, where each object is described by its category, location, size, and appearance. Details are provided in Appendix~\ref{apdx:multi_modal}.

\textbf{Quality Check} To address LLM hallucination, we implement quality control. For RelSpatialQA, human annotators verify the correctness of triplets, discarding those that are illogical or nonexistent. Grammatical errors are also corrected.
For AbsSpatialQA, annotators filter out incorrect spatial expressions and their corresponding QAs.

\textbf{Dataset Statistics}
After refinement, we obtain a total of \textbf{5,523} QA pairs in RelSpatialQA. Note that, according to Equation~\ref{eq:spatial-reasoning} and Figure~\ref{fig:cover}, a binary spatial relationship corresponds to four QA pairs in RelSpatialQA. We assume that if a model can correctly answer all four QAs, it demonstrates a complete understanding of the binary spatial relationship. We denote these four QAs as a QA quadruplet. In total, RelSpatialQA contains \textbf{1,104} quadruplets. In AbsSpatialQA, a total of \textbf{2,672} QAs are collected.


\begin{figure*}[t]
  \centering
  \includegraphics[width=0.9\linewidth]{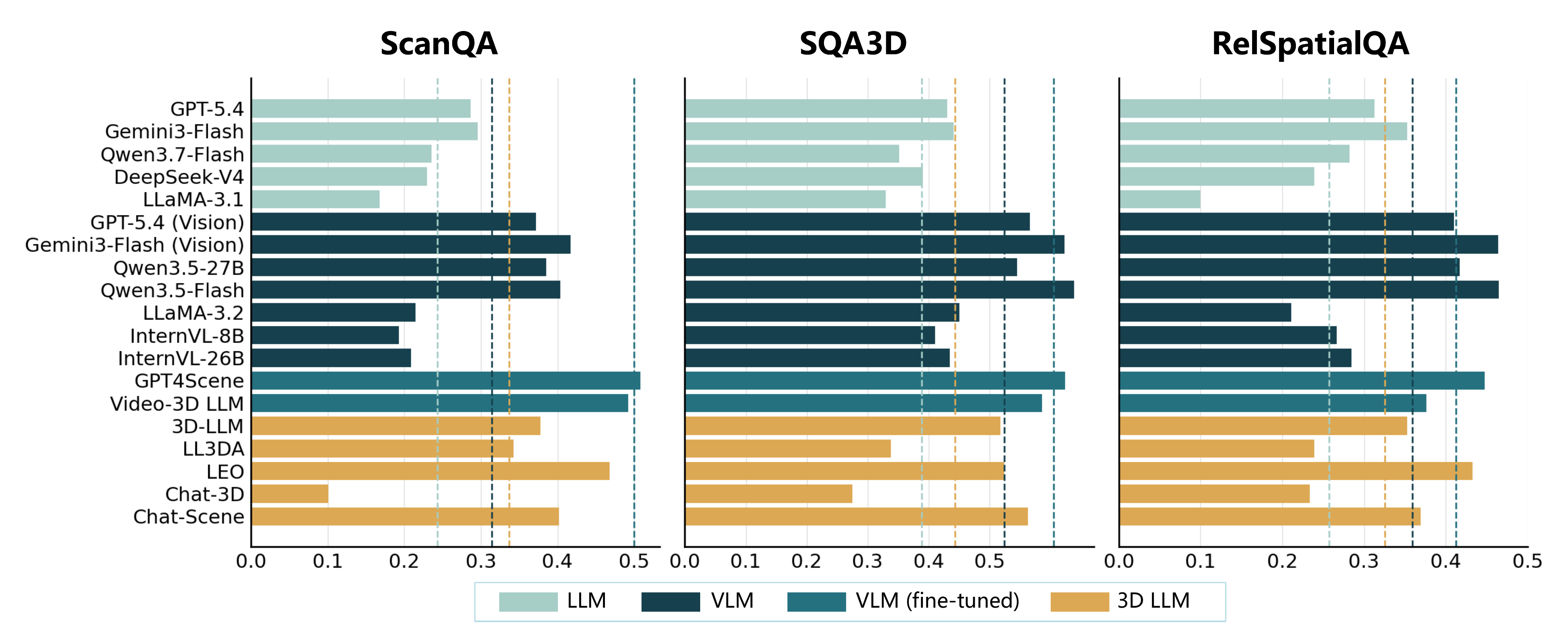}
  \vspace{-0.3cm}
  \caption{\textbf{Performance overview on ScanQA, SQA3D, and ScanReQA with different modality input.}}
  \vspace{-0.3cm}
  \label{fig:acc_multi_modal}
\end{figure*}

\section{Evaluation on Multi-modal Inputs}
To investigate the effect of different modalities, we conduct controlled experiments by altering the input modalities of MLLMs. Since scenes can be represented in text, vision, or 3D point clouds, our key idea is to replace one modality with another (e.g., substituting point clouds with images or images with text). We design three modality combinations for LLMs, VLMs, and 3D LLMs, respectively:
1) Text-only: scene descriptions as input;
2) Vision-only: multi-view images of the scene;
3) Point cloud-Vision: matches the original 3D LLM input, including point clouds with either RGB values or multi-view images.


\section{Experiments}
We first provide the details of the experimental setup and then demonstrate the following: overall performance of MLLMs, ablations of multimodal inputs, and in-depth analysis of 3D MLLM results. These correspond to addressing the following three research questions (RQs):
\begin{itemize}[leftmargin=*,partopsep=0pt,topsep=0pt]
\setlength{\itemsep}{0pt}
\setlength{\parsep}{0pt}
\setlength{\parskip}{0pt}
    \item \textbf{RQ1}: What's the performance of 3D LLMs compared to other models?
    \item \textbf{RQ2}: Do 3D LLMs truly perform reasoning?
    \item \textbf{RQ3}: How do 3D LLMs leverage point clouds to answer spatial-related questions?
\end{itemize}

\subsection{Experimental Setup}
\noindent \textbf{Models} We evaluate three different types of large models from existing SOTA works. 1) LLMs: Basic LLMs that accept only text input. We select competitive LLMs, including Qwen3.7 \cite{hui2024qwen2}, DeepSeek V4 \cite{liu2024deepseek}, and LLaMA-3 \cite{dubey2024llama}. 2) VLMs: Multi-modal LLMs that support visual input. For evaluation, we choose GPT-5.4 \cite{achiam2023gpt}, Gemini-3 Flash \cite{team2024gemini}, LLaMA-3-11B \cite{dubey2024llama}, InternVL 8B/26B \cite{chen2024internvl}, and Qwen3.5 27B/Flash \cite{bai2023qwen}. 3) 3D LLMs: 3D multi-modal LLMs that support point cloud input. We evaluate SOTA methods, including 3D-LLM \cite{hong20233d}, LL3DA \cite{chen2024ll3da}, LEO \cite{huang2023embodied}, Chat-3D v2 \cite{wang2023chat} (we term it as Chat-3D for simplicity), and Chat-Scene \cite{huang2023chat}. The inputs for the three types of models are TI, VI, and PVI, respectively. 

\noindent \textbf{Datasets} We evaluate all models on two open-source 3D QA benchmarks, ScanQA and SQA3D, as well as our proposed ScanReQA. For ScanReQA, only 3D LLMs are evaluated on the AbsSpatial QA dataset.

\noindent \textbf{Metrics} We use the refined top-1 exact match (EM@1) introduced in LEO~\cite{huang2023embodied} for overall performance comparison. For RelSpatialQA, we calculate the accuracy and recall of understanding spatial relationships, which are defined as follows:
\begin{equation}
\setlength\abovedisplayskip{3pt}
\setlength\belowdisplayskip{3pt}
\label{eq:spatial_acc}
\text{Accuracy} = \frac{N_{\text{correct}}^{\text{quad}}}{N_{\text{total}}^{\text{quad}}},\
 \text{Recall} = \frac{N_{\text{correct}}^{\text{quad}}}{N_{\text{correct}}^{\text{ScanQA}}}.
\end{equation}
where $N_{\text{correct}}^{\text{quad}}$, $N_{\text{total}}^{\text{quad}}$, $N_{\text{correct}}^{\text{ScanQA}}$ denote the number of correctly answered quadruplets, the total number of quadruplets, and the number of correctly answered QAs in ScanQA, respectively. More detailed explanations of recall are provided in Appendix~\ref{apdx:metrics}.
For AbsSpatial QA, we simply calculate the accuracy of multiple-choice questions.

\subsection{What's the performance of 3D LLMs?}
The EM@1 scores of all evaluated models are presented in Figure~\ref{fig:acc_multi_modal}, with vertical dashed lines indicating the average EM@1 for each model type.



\textbf{3D LLMs with point cloud input demonstrate stronger performance than zero-shot LLMs}
On ScanQA, SQA3D, and RelSpatialQA, 3D LLMs achieve average EM@1 scores of 33.8\%, 44.3\%, and 32.5\%, respectively. In comparison, LLMs obtain EM@1 scores of 24.3\%, 38.9\%, and 25.7\% on the same datasets, which are lower than those of 3D LLMs by 9.5\%, 5.4\%, and 6.8\%, respectively. These results indicate that, without fine-tuning, current LLMs may not reliably leverage coordinate information for spatial reasoning.


\textbf{The vision modality exhibits competitive performance compared to the point cloud modality.} 
As shown in Figure~\ref{fig:acc_multi_modal}, zero-shot VLMs obtain 31.4\% EM@1 on ScanQA which approaching the performance of 3D LLMs. The advanced proprietary models such as GPT and Gemini are even better than 3D LLMs on SQA3D and RelSpatialQA. \textbf{Fine-tuning significantly enhances the spatial reasoning ability of VLMs}, yielding the highest EM@1 in both ScanQA, SQA3D, and RelSpatialQA. These results highlight the strong potential of VLMs with multiview image inputs for 3D spatial understanding.


\begin{table}[t]
\renewcommand{\arraystretch}{1}
\small
\caption{Quantitative results on ScanReQA.}
\label{tab:res_relqa_absqa}
\centering
\vspace{-0.2cm}
\resizebox{\linewidth}{!}{
\begin{tabular}{lccc}
\toprule
\multirow{2}*{\textbf{Methods}} & \multicolumn{2}{c}{\textbf{RelSpatialQA}} & \multicolumn{1}{c}{\textbf{AbsSpatialQA}}  \\
~  & Acc. & Recall & Acc. \\ \hline
\rowcolor{blue!20} Text-only & ~ & ~ & ~ \\
GPT-5.4 & 4.0 & 7.0 & - \\
Gemini3-Flash & 4.6 & 7.9 & - \\
Qwen3.7-Flash & 4.0 & 10.0 & - \\
DeepSeek-V4 & 3.4 & 7.4 & - \\
LLaMA-3-8B & 1.6 & 4.8 & -\\ 
Average & 3.5 & 7.42 & -\\ 
\hline
\rowcolor{blue!20}Vision-only  & ~ & ~ & ~\\
GPT-5.4 & 7.3 & 13.9 & - \\
Gemini3-Flash & 9.2 & 14.9 & - \\
LLaMA-3-11B  & 2.3 & 3.2 & - \\
InternVL-8B & 2.4 & 4.1 & - \\
InternVL-26B & 3.1 & 6.7 & - \\
Qwen3.5-27B & 7.0 & 12.6 & - \\ 
Qwen3.5-Flash & 7.3 & 12.1 & - \\ 
Average & 5.5 & 10.5 & - \\ 
\hline
\rowcolor{blue!20} Vision-only (\textit{fine-tuned}) & ~ & ~ & ~\\
GPT4Scene & 12.2 & 17.5  & -  \\ 
Video-3D-LLM & 10.3 & 14.5 & -  \\
Average & 11.3 & 16.0 & - \\ 
\hline
\rowcolor{blue!20} Point cloud-vision & ~ & ~ & ~\\
3D-LLM & 5.7 & 8.7 & 24.2 \\
LL3DA & 5.3 & 9.7 & 1.5\\
LEO  & 11.2 & 12.9 & 0.3 \\
Chat-3D & 1.0 & 2.5 & 8.3 \\
Chat-Scene & 8.4 & 13.3 & 23.4 \\
Average & 6.3 & 9.4 & 5.4 \\ 
\hline
Overall Average & 8.3 & 12.7 & - \\
\bottomrule
\end{tabular}
}
\vspace{-0.5cm}
\end{table}

\subsection{Do 3D LLMs Perform Reasoning?}
To answer RQ2, we assume that the 3D LLM should be able to answer the quadruplet QAs in RelSpatialQA if it truly understands the binary spatial relationship. We also investigate whether 3D LLMs can leverage the coordinate information of the point clouds for more complex spatial reasoning. Thus, we analyze the performance of baseline models on ScanReQA.

\textbf{Elementary spatial reasoning remains challenging} We report the accuracy and recall of all models on RelSpatialQA. The results show that: 1) overall accuracy and recall are extremely low, at only 8.3\% and 12.7\%, respectively; 2) LLMs achieve the highest accuracy while zero-shot VLMs achieve the lowest, with 3D LLMs falling between them. These findings indicate that current models struggle to understand the basic binary spatial relationships for reasoning. Furthermore, incorporating point cloud information does not significantly improve the spatial relationship understanding or reasoning capabilities.

\begin{table}[t]
\centering
\small
\caption{Error breakdown under RelSpatialQA (Rel.) and AbsSpatialQA (Abs.). \textit{Re. Succ./Fail} denotes cases where the model’s response is semantically consistent/inconsistent with the ground-truth answer. \textit{Irr. Resp.} refers to responses that are unrelated to the question. \textit{Fmt. Err.} indicates responses that do not follow the required answer format.}
\label{tab:err_break}
\vspace{-0.3cm}
\resizebox{\linewidth}{!}{
\begin{tabular}{lcccccccc}
\toprule
 & \multicolumn{2}{c}{Re. Succ.}
 & \multicolumn{2}{c}{Re. Fail.}
 & \multicolumn{2}{c}{Irr. Resp.}
 & \multicolumn{2}{c}{Fmt. Err.} \\
\cmidrule(lr){2-3} \cmidrule(lr){4-5} \cmidrule(lr){6-7} \cmidrule(lr){8-9}
Dataset & Rel. & Abs. & Rel. & Abs. & Rel. & Abs. & Rel. & Abs. \\
\midrule
3D-LLM     & 5.7 & 24.2 & 94.3 & 73.1 & 0.0 & 2.7  & 0.0 & 0.0 \\
LL3DA      & 5.4 & 1.5  & 92.7 & 3.5  & 2.1 & 94.9 & 0.0 & 0.0 \\
LEO        & 11.2& 0.3  & 81.1 & 0.5  & 7.7 & 99.2 & 0.0 & 0.0 \\
Chat-3D    & 1.0 & 8.3  & 91.5 & 29.5 & 6.0 & 53.1 & 1.5 & 9.1 \\
Chat-Scene & 8.4 & 23.4 & 91.6 & 70.6 & 0.0 & 6.0  & 0.0 & 0.0 \\
\bottomrule
\end{tabular}
}
\vspace{-0.7cm}
\end{table}

\textbf{3D LLMs suffer from reasoning failure and irrelevant responses.} 
As shown in Table~\ref{tab:err_break}, we categorize model errors into reasoning failure, irrelevant response, and formatting error. On RelSpatialQA, errors of 3D LLMs are dominated by reasoning failures. Even the best-performing model, LEO, exhibits an 81.1\% reasoning failure rate, indicating that answering complete QA quadruplets remains challenging. On AbsSpatialQA, errors of 3D-LLM and Chat-Scene are mainly attributed to reasoning failures, whereas LL3DA, LEO, and Chat-3D primarily produce irrelevant responses due to changes in question format. Notably, even after excluding irrelevant responses and formatting errors, reasoning accuracy remains below 50\%.

\begin{figure}[t]
  \centering
  \includegraphics[width=0.9\linewidth]{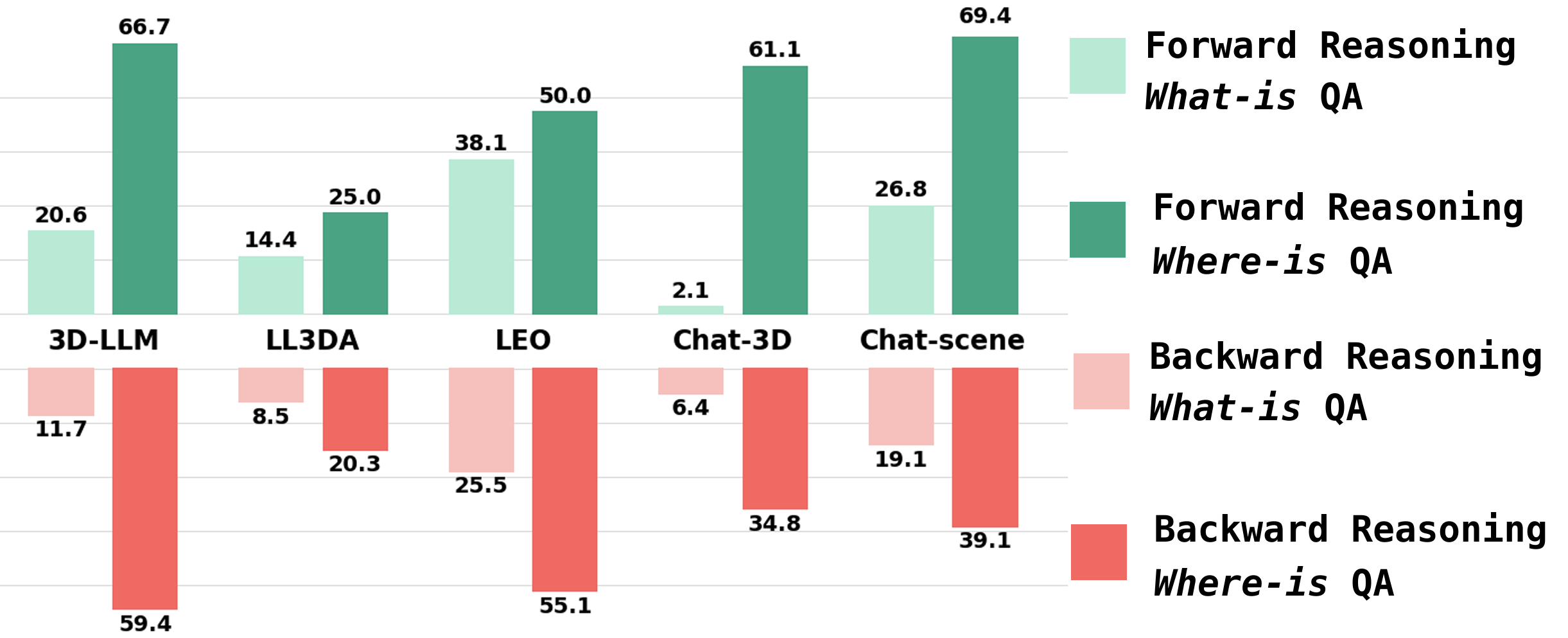}
  \vspace{-0.3cm}
  \caption{EM@1 for each QA in a QA quadruplet.}
  \vspace{-0.4cm}
  \label{fig:acc_quadruplet}
\end{figure}

\textbf{3D LLMs struggle to reason for backward spatial relationships} 
To analyze the causes of reasoning failures on QA quadruplets, we compute the accuracy of model responses to each individual question within a quadruplet, as depicted in Figure~\ref{fig:acc_quadruplet}.
We observe that model accuracy on backward spatial reasoning is substantially lower than that on forward spatial reasoning. In other words, even when 3D LLMs can correctly answer forward spatial reasoning questions, they often fail on the corresponding backward ones.
This result suggests that current 3D LLMs do not effectively leverage knowledge of forward spatial relationships to reason about their backward counterparts.

\subsection{How do 3D LLMs Leverage Point Clouds? A Representation Perspective.}
To answer this question, we conduct an in-depth analysis of the relationship between model attention to point clouds and QA accuracy, as well as interpret point cloud embeddings into the vocabulary to explore the information flow during the reasoning process. Our study focuses on LEO~\cite{huang2023embodied}, a representative open-source 3D LLM using the popular Vicuna-7B language model.

\begin{table}[t]
\vspace{-0.2em} 
\caption{Ablations on masking different tokens.}
\vspace{-0.5em} 
\centering
\small
\setlength{\tabcolsep}{2pt} 
\renewcommand{\arraystretch}{1.1}
\resizebox{\linewidth}{!}{
\begin{tabular}{@{}lcccc@{}}
\toprule
~ & w/ sink (tokens) & wo/ sink & w/ GT (tokens) & wo/ GT \\
\midrule
ScanQA     & 29.5   & 31.8  &  29.5 & 26.5     \\
RelSpatialQA    & 28.5   & 28.4  & 28.5 & 26.3  \\
\bottomrule
\end{tabular}
}
\vspace{-0.5em} 
\label{tab:sink_acc}
\vspace{-0.5em} 
\end{table}

\textbf{Analysis from the Attention Perspective}
We analyze the model's response attention to input tokens by computing the average attention scores from response tokens to point cloud tokens across all transformer layers. Details of the computation are provided in Appendix~\ref{apdx:attn_analysis}. Representative 3D LLMs such as Chat-Scene, Chat-3D, and LEO segment object-level point clouds and encode each object into a single token, which is then fed into the LLM backbone. \textit{To account for the unordered nature of point clouds, we randomly shuffle the point cloud tokens prior to input}. The key observations and findings are summarized below.

\begin{figure}[t]
  \centering
  \includegraphics[width=0.9\linewidth]{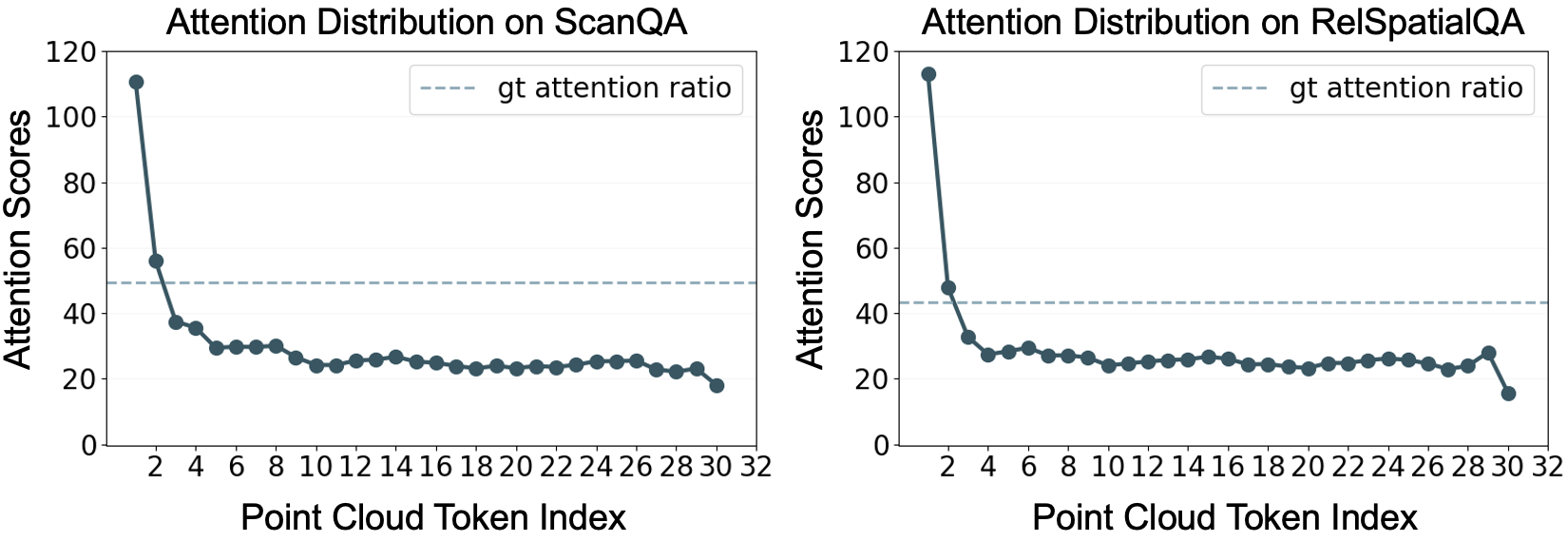}
  \vspace{-0.3cm}
  \caption{Attention distribution over token indices.}
  \vspace{-0.3cm}
  \label{fig:att_token_idx}
\end{figure}

\begin{figure}[t]
  \centering
  \includegraphics[width=0.9\linewidth]{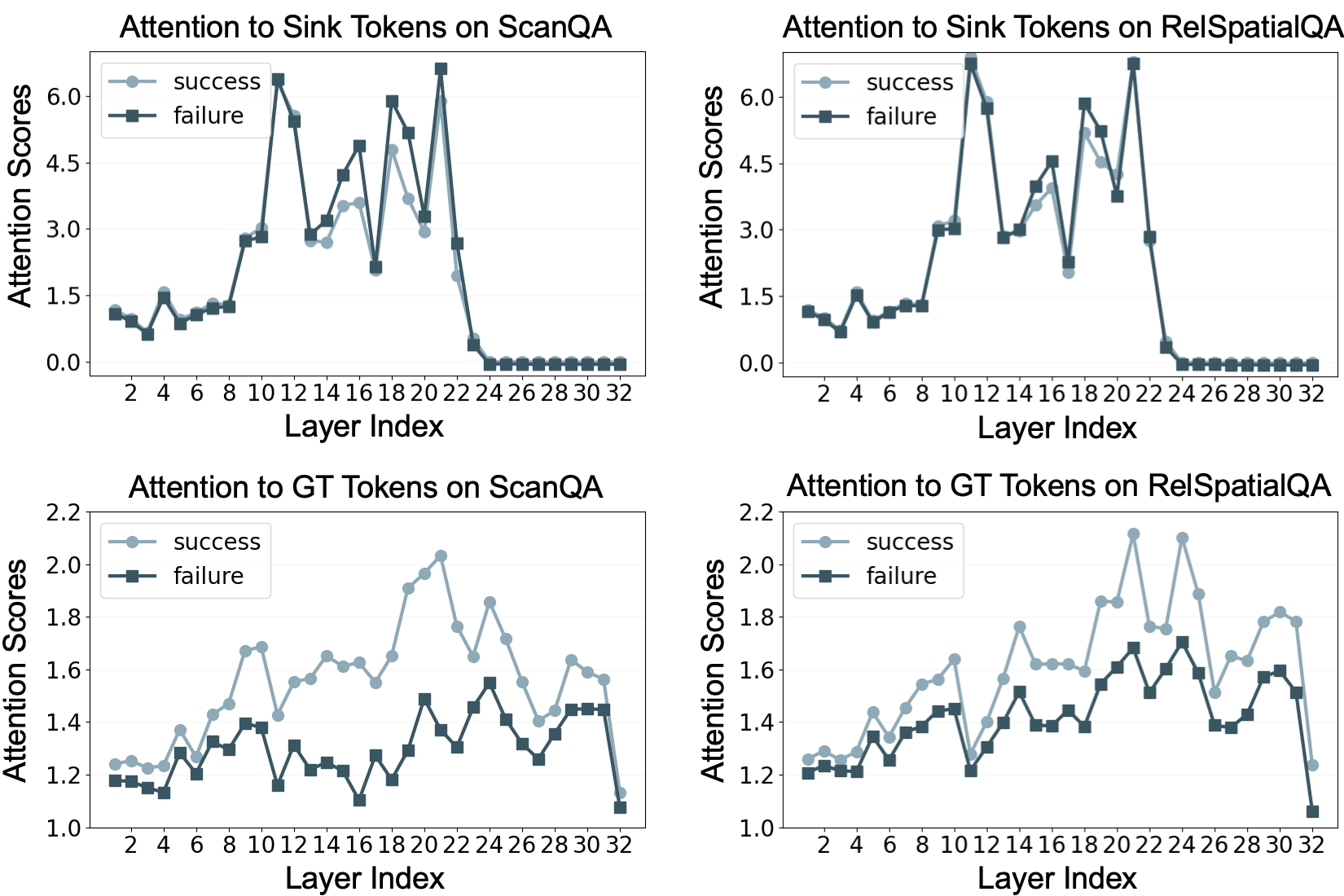}
  \caption{The attention distribution of successful and failed cases over transformer layers.}
  \vspace{-0.4cm}
  \label{fig:att_good_bad}
\end{figure}

\begin{itemize}[leftmargin=*,partopsep=0pt,topsep=0pt]
\setlength{\itemsep}{0pt}
\setlength{\parsep}{0pt}
\setlength{\parskip}{0pt}
    \item \textbf{Attention sink exists in 3D LLM} As depicted in Figure~\ref{fig:att_token_idx}, the results show that the model's responses consistently exhibit high attention to the first few object tokens. In other words, attention is concentrated on fixed token positions, which is known as the attention sink phenomenon observed in VLMs~\cite{xiao2023efficient}. We refer to these fixed-position tokens as \textbf{sink tokens}.
    
    \item \textbf{Attention on sink tokens does not contribute to reasoning} 
    We mask sink tokens in the input and evaluate the 3D LLM’s performance, as shown in Table~\ref{tab:sink_acc}. We also compare the attention scores of sink tokens between successful and failed QA cases (Figure~\ref{fig:att_good_bad}). The results show that masking sink tokens does not reduce QA accuracy and even slightly improves it on the ScanQA dataset. This suggests that sink tokens contribute little to reasoning and may even hinder the model’s reasoning process~\cite{xiao2023efficient}.

    \item \textbf{Attention on QA-Relevant Object Tokens is necessary for reasoning} We denote the object tokens correspond to anchor or target objects in the spatial relationship as \textbf{ground-truth (GT) tokens}. To assess their importance, we 1) mask these tokens and measure performance drop, and 2) compare their attention scores in successful and failed QA cases. As shown in Table~\ref{tab:sink_acc} and Figure~\ref{fig:att_good_bad}, greater attention to GT tokens correlates with higher QA accuracy, while masking them results in a performance drop of more than 2\%. This highlights their necessity for effective spatial reasoning.
\end{itemize}

\begin{figure}[t]
  \centering
  \includegraphics[width=\linewidth]{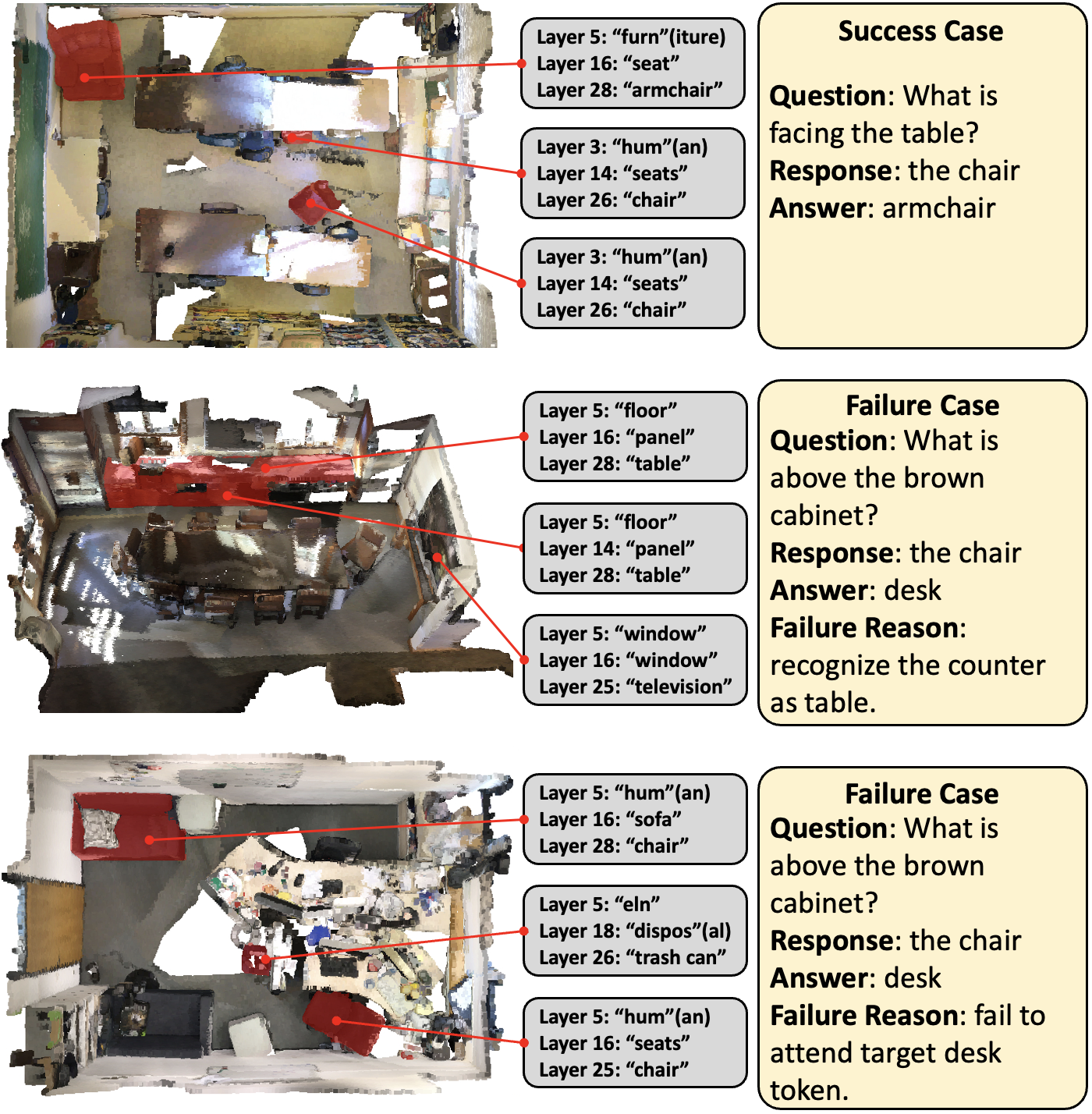}
  \caption{The attention distribution of successful and failed cases over transformer layers.}
  \vspace{-0.4cm}
  \label{fig:logit_len_ana}
\end{figure}

\textbf{Analysis from the Feature Perspective}
We use the \textit{logit lens}~\cite{nostalgebraist2020logitlens} method to see how the representations of point cloud tokens evolve through the transformer layers. The logit lens technique uses a decoder to interpret intermediate embeddings into the vocabulary. Detailed experimental settings are provided in Appendix~\ref{apdx:logit_len}.
We select QA cases whose GT tokens have the highest attention scores and depict the logit lens results in Figure~\ref{fig:logit_len_ana}. 

\begin{itemize}[leftmargin=*,partopsep=0pt,topsep=0pt]
\setlength{\itemsep}{0pt}
\setlength{\parsep}{0pt}
\setlength{\parskip}{0pt}
    \item \textbf{Point cloud representations evolve into interpretable vocabulary tokens} As shown in Figure~\ref{fig:logit_len_ana}, we find that the point cloud embeddings in the shallow and late layers correspond to tokens in the vocabulary that describe its original object in the scene. This demonstrates the applicability of the logit lens to 3D LLMs, extending its use beyond text and vision~\cite{neo2024towards} to the point cloud modality.
    \item \textbf{Incorrect token interpretation and token attention leads to reasoning failure} As shown in the second case in Figure~\ref{fig:logit_len_ana}, although the model attends to the correct tokens, it fails to align point cloud semantics with the corresponding textual vocabulary during inference, resulting in incorrect answers. In the last case, the model pays more attention on irrelevant objects, even if the model interprets the point cloud token correctly.
\end{itemize}

\textit{Discussion} The token sink phenomenon observed in 3D LLMs is consistent with that reported in 2D VLMs. One potential solution is attention redistribution~\cite{kang2025see}. It reallocates attention weights from sink tokens to non-sink tokens, allowing the model to process point cloud information more effectively. 
Object semantic misalignment between point clouds and text remains an open problem. However, this issue can be decoupled from spatial reasoning, as the spatial relationships between objects are independent of their semantic categories. Thus, by constructing appropriate spatial representations as topological graphs~\cite{ma2025spatialreasoner}, models could learn spatial concepts without being affected by object semantic misalignment.

\section{Conclusion}




Our results suggest that current 3D LLMs are still in the early stage of advancing 3D spatial reasoning. Compared with 2D LLMs, their reasoning remains less mature, particularly in aligning point cloud and textual modalities and in leveraging the quantitative geometric information. Consequently, there remains substantial room for improvement in effectively utilizing point cloud information to enhance spatial reasoning performance.


\section{Limitations}
First, this work focuses primarily on binary spatial reasoning, without extending the analysis to more complex multi-object spatial relationships. While binary relations form the foundation of higher-order spatial reasoning, it remains an open question whether current 3D LLMs exhibit similar reasoning behaviors in multi-object scenarios.

Second, we provide an in-depth analysis of the challenges faced by current 3D LLMs and discuss potential directions for improving their reasoning capabilities. However, we do not propose concrete model-level solutions for enhancing spatial reasoning. One possible direction is attention reassignment~\cite{kang2025see}, which has been explored in 2D LLMs to encourage models to attend to more informative tokens. The effectiveness of such strategies for 3D LLMs still requires further empirical validation.


{
    \small
    \bibliography{custom}

@article{ hong20233d,
  title={3d-llm: Injecting the 3d world into large language models},
  author={Hong, Yining and Zhen, Haoyu and Chen, Peihao and Zheng, Shuhong and Du, Yilun and Chen, Zhenfang and Gan, Chuang},
  journal={Advances in Neural Information Processing Systems},
  volume={36},
  pages={20482--20494},
  year={2023}
}

@article{wang2023chat,
  title={Chat-3d: Data-efficiently tuning large language model for universal dialogue of 3d scenes},
  author={Wang, Zehan and Huang, Haifeng and Zhao, Yang and Zhang, Ziang and Zhao, Zhou},
  journal={arXiv preprint arXiv:2308.08769},
  year={2023}
}

@article{huang2023embodied,
  title={An embodied generalist agent in 3d world},
  author={Huang, Jiangyong and Yong, Silong and Ma, Xiaojian and Linghu, Xiongkun and Li, Puhao and Wang, Yan and Li, Qing and Zhu, Song-Chun and Jia, Baoxiong and Huang, Siyuan},
  journal={arXiv preprint arXiv:2311.12871},
  year={2023}
}

@inproceedings{chen2024ll3da,
  title={LL3DA: Visual Interactive Instruction Tuning for Omni-3D Understanding Reasoning and Planning},
  author={Chen, Sijin and Chen, Xin and Zhang, Chi and Li, Mingsheng and Yu, Gang and Fei, Hao and Zhu, Hongyuan and Fan, Jiayuan and Chen, Tao},
  booktitle={Proceedings of the IEEE/CVF Conference on Computer Vision and Pattern Recognition},
  pages={26428--26438},
  year={2024}
}

@inproceedings{hong2024multiply,
  title={Multiply: A multisensory object-centric embodied large language model in 3d world},
  author={Hong, Yining and Zheng, Zishuo and Chen, Peihao and Wang, Yian and Li, Junyan and Gan, Chuang},
  booktitle={Proceedings of the IEEE/CVF Conference on Computer Vision and Pattern Recognition},
  pages={26406--26416},
  year={2024}
}

@article{alayrac2022flamingo,
  title={Flamingo: a visual language model for few-shot learning},
  author={Alayrac, Jean-Baptiste and Donahue, Jeff and Luc, Pauline and Miech, Antoine and Barr, Iain and Hasson, Yana and Lenc, Karel and Mensch, Arthur and Millican, Katherine and Reynolds, Malcolm and others},
  journal={Advances in neural information processing systems},
  volume={35},
  pages={23716--23736},
  year={2022}
}

@article{driess2023palm,
  title={Palm-e: An embodied multimodal language model},
  author={Driess, Danny and Xia, Fei and Sajjadi, Mehdi SM and Lynch, Corey and Chowdhery, Aakanksha and Ichter, Brian and Wahid, Ayzaan and Tompson, Jonathan and Vuong, Quan and Yu, Tianhe and others},
  journal={arXiv preprint arXiv:2303.03378},
  year={2023}
}

@article{liu2024visual,
  title={Visual instruction tuning},
  author={Liu, Haotian and Li, Chunyuan and Wu, Qingyang and Lee, Yong Jae},
  journal={Advances in neural information processing systems},
  volume={36},
  year={2024}
}

@article{rana2023sayplan,
  title={Sayplan: Grounding large language models using 3d scene graphs for scalable task planning},
  author={Rana, Krishan and Haviland, Jesse and Garg, Sourav and Abou-Chakra, Jad and Reid, Ian D and Suenderhauf, Niko},
  journal={CoRR},
  year={2023}
}

@inproceedings{zheng2024towards,
  title={Towards learning a generalist model for embodied navigation},
  author={Zheng, Duo and Huang, Shijia and Zhao, Lin and Zhong, Yiwu and Wang, Liwei},
  booktitle={Proceedings of the IEEE/CVF Conference on Computer Vision and Pattern Recognition},
  pages={13624--13634},
  year={2024}
}

@article{mirjalili2023lan,
  title={Lan-grasp: Using large language models for semantic object grasping},
  author={Mirjalili, Reihaneh and Krawez, Michael and Silenzi, Simone and Blei, Yannik and Burgard, Wolfram},
  journal={arXiv preprint arXiv:2310.05239},
  year={2023}
}

@inproceedings{azuma2022scanqa,
  title={Scanqa: 3d question answering for spatial scene understanding},
  author={Azuma, Daichi and Miyanishi, Taiki and Kurita, Shuhei and Kawanabe, Motoaki},
  booktitle={proceedings of the IEEE/CVF conference on computer vision and pattern recognition},
  pages={19129--19139},
  year={2022}
}

@article{ma2022sqa3d,
  title={Sqa3d: Situated question answering in 3d scenes},
  author={Ma, Xiaojian and Yong, Silong and Zheng, Zilong and Li, Qing and Liang, Yitao and Zhu, Song-Chun and Huang, Siyuan},
  journal={arXiv preprint arXiv:2210.07474},
  year={2022}
}

@inproceedings{dai2017scannet,
  title={Scannet: Richly-annotated 3d reconstructions of indoor scenes},
  author={Dai, Angela and Chang, Angel X and Savva, Manolis and Halber, Maciej and Funkhouser, Thomas and Nie{\ss}ner, Matthias},
  booktitle={Proceedings of the IEEE conference on computer vision and pattern recognition},
  pages={5828--5839},
  year={2017}
}

@inproceedings{jia2025sceneverse,
  title={Sceneverse: Scaling 3d vision-language learning for grounded scene understanding},
  author={Jia, Baoxiong and Chen, Yixin and Yu, Huangyue and Wang, Yan and Niu, Xuesong and Liu, Tengyu and Li, Qing and Huang, Siyuan},
  booktitle={European Conference on Computer Vision},
  pages={289--310},
  year={2025},
  organization={Springer}
}

@inproceedings{zhang2023multi3drefer,
  title={Multi3drefer: Grounding text description to multiple 3d objects},
  author={Zhang, Yiming and Gong, ZeMing and Chang, Angel X},
  booktitle={Proceedings of the IEEE/CVF International Conference on Computer Vision},
  pages={15225--15236},
  year={2023}
}

@article{achiam2023gpt,
  title={Gpt-4 technical report},
  author={Achiam, Josh and Adler, Steven and Agarwal, Sandhini and Ahmad, Lama and Akkaya, Ilge and Aleman, Florencia Leoni and Almeida, Diogo and Altenschmidt, Janko and Altman, Sam and Anadkat, Shyamal and others},
  journal={arXiv preprint arXiv:2303.08774},
  year={2023}
}

@article{zhu2024llava,
  title={Llava-3d: A simple yet effective pathway to empowering lmms with 3d-awareness},
  author={Zhu, Chenming and Wang, Tai and Zhang, Wenwei and Pang, Jiangmiao and Liu, Xihui},
  journal={arXiv preprint arXiv:2409.18125},
  year={2024}
}

@article{touvron2023llama,
  title={Llama: Open and efficient foundation language models},
  author={Touvron, Hugo and Lavril, Thibaut and Izacard, Gautier and Martinet, Xavier and Lachaux, Marie-Anne and Lacroix, Timoth{\'e}e and Rozi{\`e}re, Baptiste and Goyal, Naman and Hambro, Eric and Azhar, Faisal and others},
  journal={arXiv preprint arXiv:2302.13971},
  year={2023}
}

@article{brown2020language,
  title={Language models are few-shot learners},
  author={Brown, Tom and Mann, Benjamin and Ryder, Nick and Subbiah, Melanie and Kaplan, Jared D and Dhariwal, Prafulla and Neelakantan, Arvind and Shyam, Pranav and Sastry, Girish and Askell, Amanda and others},
  journal={Advances in neural information processing systems},
  volume={33},
  pages={1877--1901},
  year={2020}
}

@article{team2023gemini,
  title={Gemini: a family of highly capable multimodal models},
  author={Team, Gemini and Anil, Rohan and Borgeaud, Sebastian and Alayrac, Jean-Baptiste and Yu, Jiahui and Soricut, Radu and Schalkwyk, Johan and Dai, Andrew M and Hauth, Anja and Millican, Katie and others},
  journal={arXiv preprint arXiv:2312.11805},
  year={2023}
}

@inproceedings{zhu20233d,
  title={3d-vista: Pre-trained transformer for 3d vision and text alignment},
  author={Zhu, Ziyu and Ma, Xiaojian and Chen, Yixin and Deng, Zhidong and Huang, Siyuan and Li, Qing},
  booktitle={Proceedings of the IEEE/CVF International Conference on Computer Vision},
  pages={2911--2921},
  year={2023}
}

@article{man2024lexicon3d,
  title={Lexicon3d: Probing visual foundation models for complex 3d scene understanding},
  author={Man, Yunze and Zheng, Shuhong and Bao, Zhipeng and Hebert, Martial and Gui, Liang-Yan and Wang, Yu-Xiong},
  journal={arXiv preprint arXiv:2409.03757},
  year={2024}
}

@article{yang2024thinking,
  title={Thinking in space: How multimodal large language models see, remember, and recall spaces},
  author={Yang, Jihan and Yang, Shusheng and Gupta, Anjali W and Han, Rilyn and Fei-Fei, Li and Xie, Saining},
  journal={arXiv preprint arXiv:2412.14171},
  year={2024}
}

@inproceedings{wald2019rio,
  title={Rio: 3d object instance re-localization in changing indoor environments},
  author={Wald, Johanna and Avetisyan, Armen and Navab, Nassir and Tombari, Federico and Nie{\ss}ner, Matthias},
  booktitle={Proceedings of the IEEE/CVF International Conference on Computer Vision},
  pages={7658--7667},
  year={2019}
}

@inproceedings{li2023blip,
  title={Blip-2: Bootstrapping language-image pre-training with frozen image encoders and large language models},
  author={Li, Junnan and Li, Dongxu and Savarese, Silvio and Hoi, Steven},
  booktitle={International conference on machine learning},
  pages={19730--19742},
  year={2023},
  organization={PMLR}
}

@article{hui2024qwen2,
  title={Qwen2. 5-coder technical report},
  author={Hui, Binyuan and Yang, Jian and Cui, Zeyu and Yang, Jiaxi and Liu, Dayiheng and Zhang, Lei and Liu, Tianyu and Zhang, Jiajun and Yu, Bowen and Lu, Keming and others},
  journal={arXiv preprint arXiv:2409.12186},
  year={2024}
}

@article{liu2024deepseek,
  title={Deepseek-v3 technical report},
  author={Liu, Aixin and Feng, Bei and Xue, Bing and Wang, Bingxuan and Wu, Bochao and Lu, Chengda and Zhao, Chenggang and Deng, Chengqi and Zhang, Chenyu and Ruan, Chong and others},
  journal={arXiv preprint arXiv:2412.19437},
  year={2024}
}

@article{dubey2024llama,
  title={The llama 3 herd of models},
  author={Dubey, Abhimanyu and Jauhri, Abhinav and Pandey, Abhinav and Kadian, Abhishek and Al-Dahle, Ahmad and Letman, Aiesha and Mathur, Akhil and Schelten, Alan and Yang, Amy and Fan, Angela and others},
  journal={arXiv preprint arXiv:2407.21783},
  year={2024}
}

@article{team2024gemini,
  title={Gemini 1.5: Unlocking multimodal understanding across millions of tokens of context},
  author={Team, Gemini and Georgiev, Petko and Lei, Ving Ian and Burnell, Ryan and Bai, Libin and Gulati, Anmol and Tanzer, Garrett and Vincent, Damien and Pan, Zhufeng and Wang, Shibo and others},
  journal={arXiv preprint arXiv:2403.05530},
  year={2024}
}

@inproceedings{chen2024internvl,
  title={Internvl: Scaling up vision foundation models and aligning for generic visual-linguistic tasks},
  author={Chen, Zhe and Wu, Jiannan and Wang, Wenhai and Su, Weijie and Chen, Guo and Xing, Sen and Zhong, Muyan and Zhang, Qinglong and Zhu, Xizhou and Lu, Lewei and others},
  booktitle={Proceedings of the IEEE/CVF conference on computer vision and pattern recognition},
  pages={24185--24198},
  year={2024}
}

@article{bai2023qwen,
  title={Qwen-VL: A Versatile Vision-Language Model for Understanding, Localization},
  author={Bai, Jinze and Bai, Shuai and Yang, Shusheng and Wang, Shijie and Tan, Sinan and Wang, Peng and Lin, Junyang and Zhou, Chang and Zhou, Jingren},
  journal={Text Reading, and Beyond},
  volume={2},
  year={2023}
}

@article{huang2023chat,
  title={Chat-Scene: Bridging 3D Scene and Large Language Models with Object Identifiers},
  author={Huang, Haifeng and Chen, Yilun and Wang, Zehan and Huang, Rongjie and Xu, Runsen and Wang, Tai and Liu, Luping and Cheng, Xize and Zhao, Yang and Pang, Jiangmiao and others},
  journal={arXiv preprint arXiv:2312.08168},
  year={2023}
}

@article{davis2023benchmarks,
  title={Benchmarks for automated commonsense reasoning: A survey},
  author={Davis, Ernest},
  journal={ACM Computing Surveys},
  volume={56},
  number={4},
  pages={1--41},
  year={2023},
  publisher={ACM New York, NY, USA}
}

@article{liu2025matheval,
  title={MathEval: A comprehensive benchmark for evaluating large language models on mathematical reasoning capabilities},
  author={Liu, Tianqiao and Chen, Zui and Fang, Zhensheng and Luo, Weiqi and Tian, Mi and Liu, Zitao},
  journal={Frontiers of Digital Education},
  volume={2},
  number={2},
  pages={16},
  year={2025},
  publisher={Springer}
}

@article{lu2022learn,
  title={Learn to explain: Multimodal reasoning via thought chains for science question answering},
  author={Lu, Pan and Mishra, Swaroop and Xia, Tanglin and Qiu, Liang and Chang, Kai-Wei and Zhu, Song-Chun and Tafjord, Oyvind and Clark, Peter and Kalyan, Ashwin},
  journal={Advances in Neural Information Processing Systems},
  volume={35},
  pages={2507--2521},
  year={2022}
}

@article{wang2025picture,
  title={Is a picture worth a thousand words? delving into spatial reasoning for vision language models},
  author={Wang, Jiayu and Ming, Yifei and Shi, Zhenmei and Vineet, Vibhav and Wang, Xin and Li, Sharon and Joshi, Neel},
  journal={Advances in Neural Information Processing Systems},
  volume={37},
  pages={75392--75421},
  year={2025}
}

@article{zhang2025llava,
  title={LLaVA-Mini: Efficient Image and Video Large Multimodal Models with One Vision Token},
  author={Zhang, Shaolei and Fang, Qingkai and Yang, Zhe and Feng, Yang},
  journal={arXiv preprint arXiv:2501.03895},
  year={2025}
}

@inproceedings{liu2024mmbench,
  title={Mmbench: Is your multi-modal model an all-around player?},
  author={Liu, Yuan and Duan, Haodong and Zhang, Yuanhan and Li, Bo and Zhang, Songyang and Zhao, Wangbo and Yuan, Yike and Wang, Jiaqi and He, Conghui and Liu, Ziwei and others},
  booktitle={European conference on computer vision},
  pages={216--233},
  year={2024},
  organization={Springer}
}

@article{hendrycks2020measuring,
  title={Measuring massive multitask language understanding},
  author={Hendrycks, Dan and Burns, Collin and Basart, Steven and Zou, Andy and Mazeika, Mantas and Song, Dawn and Steinhardt, Jacob},
  journal={arXiv preprint arXiv:2009.03300},
  year={2020}
}

@inproceedings{man2024situational,
  title={Situational awareness matters in 3d vision language reasoning},
  author={Man, Yunze and Gui, Liang-Yan and Wang, Yu-Xiong},
  booktitle={Proceedings of the IEEE/CVF Conference on Computer Vision and Pattern Recognition},
  pages={13678--13688},
  year={2024}
}

@inproceedings{zhi2025lscenellm,
  title={Lscenellm: Enhancing large 3d scene understanding using adaptive visual preferences},
  author={Zhi, Hongyan and Chen, Peihao and Li, Junyan and Ma, Shuailei and Sun, Xinyu and Xiang, Tianhang and Lei, Yinjie and Tan, Mingkui and Gan, Chuang},
  booktitle={Proceedings of the Computer Vision and Pattern Recognition Conference},
  pages={3761--3771},
  year={2025}
}

@article{peng2025understanding,
  title={Understanding and evaluating hallucinations in 3d visual language models},
  author={Peng, Ruiying and Li, Kaiyuan and Zhang, Weichen and Gao, Chen and Chen, Xinlei and Li, Yong},
  journal={arXiv preprint arXiv:2502.15888},
  year={2025}
}

@inproceedings{zheng2025video,
  title={Video-3d llm: Learning position-aware video representation for 3d scene understanding},
  author={Zheng, Duo and Huang, Shijia and Wang, Liwei},
  booktitle={Proceedings of the Computer Vision and Pattern Recognition Conference},
  pages={8995--9006},
  year={2025}
}

@article{xiao2023efficient,
  title={Efficient streaming language models with attention sinks},
  author={Xiao, Guangxuan and Tian, Yuandong and Chen, Beidi and Han, Song and Lewis, Mike},
  journal={arXiv preprint arXiv:2309.17453},
  year={2023}
}

@misc{nostalgebraist2020logitlens,
  author       = {Nostalgebraist},
  title        = {Interpreting GPT: The Logit Lens},
  howpublished = {\url{https://www.alignmentforum.org/posts/AcKRB8wDpdaN6v6ru/interpreting-gpt-the-logit-lens}},
  year         = {2020},
  month        = {August},
  note         = {Accessed: 2024-09-23}
}

@article{neo2024towards,
  title={Towards interpreting visual information processing in vision-language models},
  author={Neo, Clement and Ong, Luke and Torr, Philip and Geva, Mor and Krueger, David and Barez, Fazl},
  journal={arXiv preprint arXiv:2410.07149},
  year={2024}
}

@article{qi2025gpt4scene,
  title={Gpt4scene: Understand 3d scenes from videos with vision-language models},
  author={Qi, Zhangyang and Zhang, Zhixiong and Fang, Ye and Wang, Jiaqi and Zhao, Hengshuang},
  journal={arXiv preprint arXiv:2501.01428},
  year={2025}
}

@article{zhang2025mllms,
  title={Mllms know where to look: Training-free perception of small visual details with multimodal llms},
  author={Zhang, Jiarui and Khayatkhoei, Mahyar and Chhikara, Prateek and Ilievski, Filip},
  journal={arXiv preprint arXiv:2502.17422},
  year={2025}
}

@article{kang2025see,
  title={See what you are told: Visual attention sink in large multimodal models},
  author={Kang, Seil and others},
  journal={arXiv:2503.03321},
  year={2025}
}

@article{linghu2024multi,
  title={Multi-modal situated reasoning in 3d scenes},
  author={Linghu, Xiongkun and Huang, Jiangyong and Niu, Xuesong and Ma, Xiaojian Shawn and Jia, Baoxiong and Huang, Siyuan},
  journal={Advances in Neural Information Processing Systems},
  volume={37},
  pages={140903--140936},
  year={2024}
}

@inproceedings{yang20253d,
  title={3d-grand: A million-scale dataset for 3d-llms with better grounding and less hallucination},
  author={Yang, Jianing and Chen, Xuweiyi and Madaan, Nikhil and Iyengar, Madhavan and Qian, Shengyi and Fouhey, David F and Chai, Joyce},
  booktitle={Proceedings of the Computer Vision and Pattern Recognition Conference},
  pages={29501--29512},
  year={2025}
}

@article{lyu2024mmscan,
  title={Mmscan: A multi-modal 3d scene dataset with hierarchical grounded language annotations},
  author={Lyu, Ruiyuan and Lin, Jingli and Wang, Tai and Yang, Shuai and Mao, Xiaohan and Chen, Yilun and Xu, Runsen and Huang, Haifeng and Zhu, Chenming and Lin, Dahua and others},
  journal={Advances in Neural Information Processing Systems},
  volume={37},
  pages={50898--50924},
  year={2024}
}

@inproceedings{huang2025unveiling,
  title={Unveiling the mist over 3d vision-language understanding: Object-centric evaluation with chain-of-analysis},
  author={Huang, Jiangyong and Jia, Baoxiong and Wang, Yan and Zhu, Ziyu and Linghu, Xiongkun and Li, Qing and Zhu, Song-Chun and Huang, Siyuan},
  booktitle={Proceedings of the Computer Vision and Pattern Recognition Conference},
  pages={24570--24581},
  year={2025}
}

@article{ma2025spatialreasoner,
  title={Spatialreasoner: Towards explicit and generalizable 3d spatial reasoning},
  author={Ma, Wufei and Chou, Yu-Cheng and Liu, Qihao and Wang, Xingrui and de Melo, Celso and Xie, Jianwen and Yuille, Alan},
  journal={arXiv preprint arXiv:2504.20024},
  year={2025}
}

@article{ma20263d,
  title={Do 3D Large Language Models Really Understand 3D Spatial Relationships?},
  author={Ma, Xianzheng and Sun, Tao and Chen, Shuai and Bhalgat, Yash and Gu, Jindong and Chang, Angel X and Armeni, Iro and Laina, Iro and Peng, Songyou and Prisacariu, Victor Adrian},
  journal={arXiv preprint arXiv:2603.23523},
  year={2026}
}
}
\newpage
\appendix
\section{Appendix}
\subsection{Multimodal Inputs Conversion}
\label{apdx:multi_modal}
\begin{figure*}[t]
  \centering
  \includegraphics[width=0.9\linewidth]{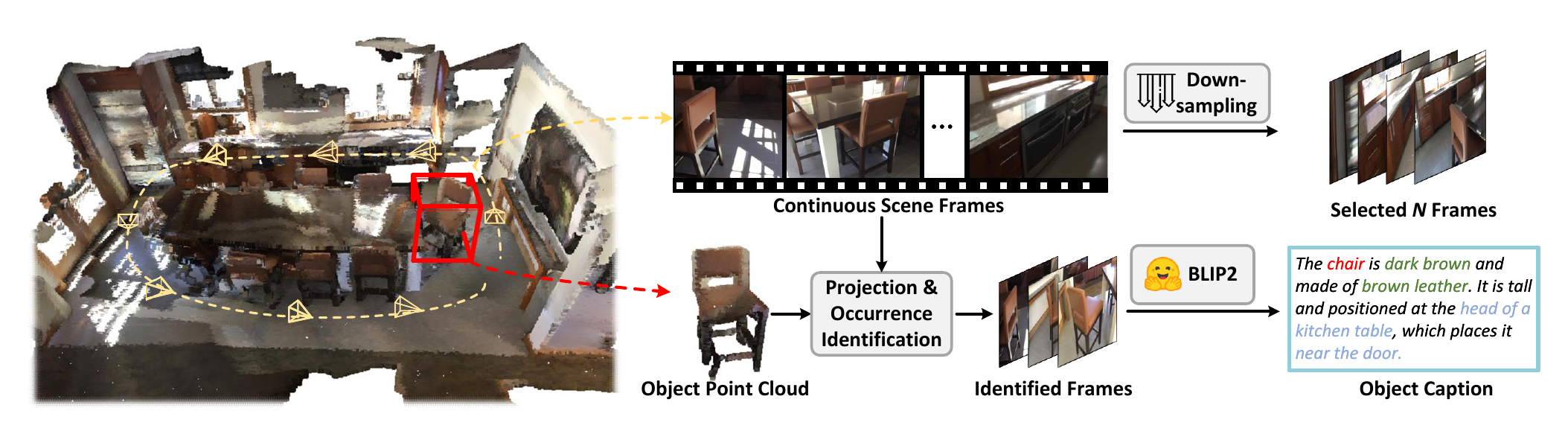}
   \vspace{-0.3cm}
  \caption{\textbf{The multi-modal data generation pipeline.} The 3D scene scan is projected into continuous scene frames which are further uniformly downsampled to $N$ frames as the input for VLMs. We also select frames containing scene objects with object point clouds and leverage caption models to generate a text description of the scene as the input for LLMs.}
  \vspace{-0.4cm}
  \label{fig:data_gen}
\end{figure*}

To ensure fair evaluation in different combinations of modality, we design a modality transformation pipeline that generates $\text{point cloud}, \text{RGB}, and \text{text}$ inputs while preserving key semantic information and minimizing information loss.

For $\textit{point cloud}$ generation, we project original multi-view RGB scans into 3D point clouds using pre-recorded camera poses and aligned depth images from 3D benchmarks like ScanNet \cite{dai2017scannet} and 3RScan \cite{wald2019rio}. Since point clouds originate from multi-view scans, they share the same semantics except for depth.

For \textit{visual} input generation, we uniformly downsample the original RGB scans to 64 frames to reduce redundancy, which is consistent with the input format of the current video LLM~\cite{zheng2025video, qi2025gpt4scene, chen2024internvl}.  Besides, Yang \textit{et al}. \cite{yang2024thinking} suggests sequential visual inputs encode spatial semantics, aiding spatial reasoning. We use these RGB frames as the point cloud counterpart.

For $\textit{text}$ input generation, we follow SceneVerse \cite{jia2025sceneverse} and generate object descriptions via image captioning. We first project object point clouds onto the 2D plane to identify their presence in RGB frames. Then, BLIP2 \cite{li2023blip} generates object captions. Since BLIP2 may leak the spatial information it directly gives the spatial relationships, we follow MSQA~\cite{linghu2024multi} which describes a scene by a collection of objects. We maintain each object's category, location, size, and appearance, which are refined into structured descriptions for prompting LLMs.

\subsection{Attention Analysis}
\label{apdx:attn_analysis}
We adopt the attention measurement method in~\cite{zhang2025llava} shown in Eq.~\ref{eq:attn1}. The input prompt to the LLM contains different types of tokens, denoted as $T_{sys}$, $T_{point\_cloud}$, $T_{instruction}$, and $T_{response}$, respectively. We quantify the attention between different tokens based on the attention map of the transformer layer in the LLM. Given an input token sequence of length $L$, each transformer layer outputs $N$ attention maps $S_i$ with shape $L \times L$ for $N$ attention heads. The $N$ attention maps are averaged across all heads to obtain an average attention map $S$. The attention weight from the 
$i^{th}$ token $t_i$ to the $j^{th}$ token $t_j$ is denoted as $S_{ij}$. For any two types of tokens $T_A$ and $T_B \in \{T_{sys}, T_{point\_cloud}, T_{instruction}, T_{response}\}$, the attention weight from $T_A$ to $T_B$ is given by:

\begin{equation}
\small
\label{eq:attn1}
    \text{Attn}(A \to B) = {
        \sum_{t_i \in T_A} \sum_{t_j \in T_B} S_{ij}
    }
\end{equation}

Particularly, $\sum_{t_i \in T_A} \sum_{t_j \in T_B} S_{ij}$ sums the all attention weights from $T_A$ to $T_B$ and $\sum_{t_i \in T_A} \mathbf{1}_{\sum_{t_j \in T_B} S_{ij} > 0}$ computes the counts of non-zero attention weight. Thus, \text{Attn} represents the average attention score from token $A$ to token $B$. In this work, we focus on the attention from response tokens to point cloud tokens.

\textbf{Relative Attention}
Due to the substantial variation in vanilla attention scores across different cases, directly averaging them may obscure case-specific attention patterns, leading to a distorted overall distribution that fails to accurately reflect the model’s true attention behavior during spatial reasoning. Thus, we adopt \textit{relative attention}~\cite{zhang2025mllms} that computes the ratio of absolute attention scores to the per-layer average scores. We define $\text{Attn}^{(k)}(A\to B_i)$ as the attention score from $T_A$ to $j$-th the token of $T_B$ in the $k$-th layer, and $\text{Attn}^{(k)}(A\to \overline{B})$ as the average attention score from $T_A$ to $T_B$ in the $k$-th layer. The relative attention score is defined as:

\begin{equation}
\small
\label{eq:attn_rel}
    \text{Attn}^{(k)}_{rel}(A \to B_j) = \frac{\text{Attn}^{(k)}(A\to B_j)}{\text{Attn}^{(k)}(A\to \overline{B})},
\end{equation}

\begin{equation}
\small
\label{eq:attn_B_i}
    \text{Attn}^{(k)}(A \to B_j) = {
        \sum_{t_i \in T_A} S_{ij}
    },
\end{equation}

\begin{equation}
\small
\label{eq:attn_avg}
    \text{Attn}^{(k)}(A \to \overline{B}) = \frac{1}{N_B}{
        \sum_{t_i \in T_A} \sum_{t_j \in T_B} S_{ij}
    },
\end{equation}
where $N_B$ is the token number of $T_B$.

\begin{figure}[t]
  \centering
    \includegraphics[width=\linewidth]{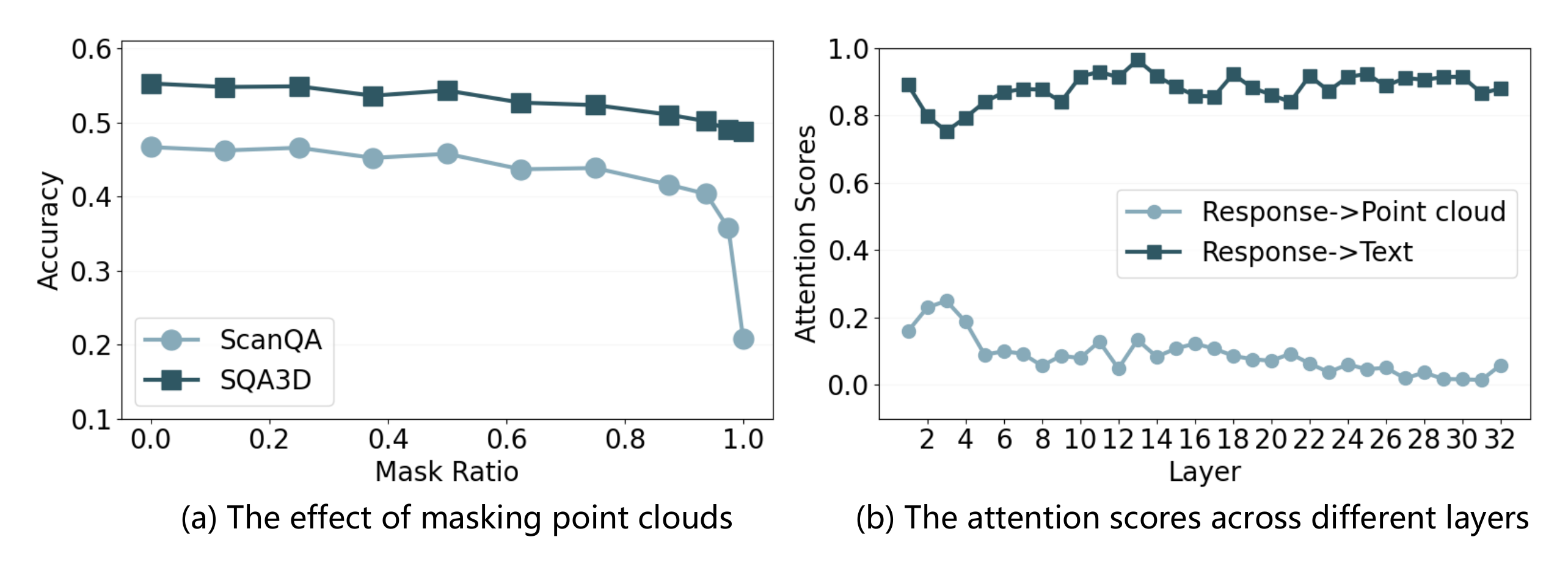}
  \caption{\textbf{Visualization of accuracy and attentions scores of LEO.} Left: the results of masking different ratios of point cloud tokens. Right: the attention scores on text and point cloud tokens.}
  \label{fig:acc_mask_pc}
\end{figure}

\textbf{Attention on Multimodal Inputs} As depicted in Figure~\ref{fig:acc_mask_pc}, we further analyze the impact of masking different proportions of point cloud tokens on spatial reasoning accuracy (left), and compare the response attention across different input modalities (right). The results show that for ScanQA, masking 90\% of point cloud tokens does not significantly affect performance. However, when only 10\% of the tokens remain, accuracy drops sharply by 24\%, indicating that only a small subset of point cloud tokens plays a critical role in spatial understanding. Additionally, Figure~\ref{fig:acc_mask_pc}(b) shows that the model's response exhibits significantly higher attention to textual inputs than to point cloud tokens, which aligns with similar findings in 2D LLMs~\cite{zhang2025llava}.

\subsection{Logit Lens for 3D LLMs}
\label{apdx:logit_len}
The logit lens is a mechanistic interpretability technique designed to understand what large language models "know" at intermediate layers. It interprets the model’s internal representations by translating hidden states back into the model’s output vocabulary space which is human-interpretable.

\begin{figure*}[t]
  \centering
    \includegraphics[width=\linewidth]{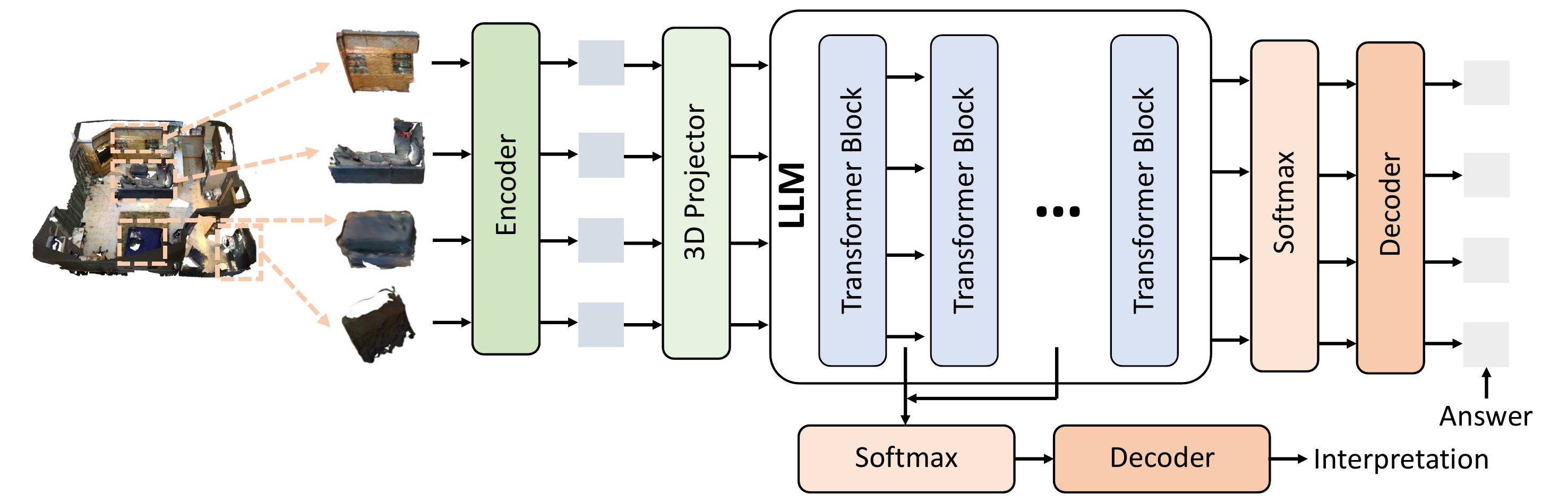}
  \caption{\textbf{The mechanism of logit lens}. The point cloud tokens are first encoded by a pretrained point cloud encoder and projected into token space via a 3D projector. After inference in the LLM backbone, the features are passed through a softmax layer and then decoded into the final answer. For simplicity, the text and visual inputs are omitted.}
  \label{fig:logit_lens}
\end{figure*}

We extend the technique to 3D LLMs as depicted in Figure~\ref{fig:logit_lens}. In a 3D LLM, each layer produces a hidden representation of the current token or sequence. Normally, only the final layer is passed through the output projection and softmax to produce logits which then determine the next predicted word. The logit lens technique applies this to earlier layers by applying the same output projection matrix to intermediate activations. This effectively shows ``what the model would predict if it stopped thinking here.''

Formally, the logits of $l$-th layer is formulates as:
\begin{equation}
    \text{logits}_l = W_U\cdot h_l,
\end{equation}
where $h_l$ is the hidden states at layer $l$ and $W_U$ is the output projection matrix. By examining these layer-wise logits,  we can visualize how predictions evolve as the model processes and refines information.


\subsection{Evaluation Metrics}
\label{apdx:metrics}
As illustrated in Figure~\ref{fig:qa_gen}, one QA in ScanQA corresponds to a spatial relationship that generates a quadruplet of QAs in RelSpatial QA, and we consider the model to understand the spatial relationship if it correctly answers all QAs in the quadruplet. Therefore, the accuracy of understanding spatial relationships is defined as follows:
\begin{equation}
\setlength\abovedisplayskip{3pt}
\setlength\belowdisplayskip{3pt}
\text{Accuracy} = \frac{N_{\text{correct}}^{\text{quad}}}{N_{\text{total}}^{\text{quad}}},\
\end{equation}
Additionally, since the QA pairs in RelSpatialQA and ScanQA partially overlap, we assume that if a model truly understands a specific spatial relationship, it should be able to correctly answer the corresponding questions in both datasets. Therefore, we define the recall metric as follows to evaluate whether the model genuinely understands the spatial relationships:
\begin{equation}
\setlength\abovedisplayskip{3pt}
\setlength\belowdisplayskip{3pt}
\label{eq:spatial_recall}
 \text{Recall} = \frac{N_{\text{correct}}^{\text{quad}}}{N_{\text{correct}}^{\text{ScanQA}}}.
\end{equation}

\subsection{Implementation Details}
\label{apdx:Implementation}
We fix the input frame count to 64 for VLMs. Specifically, we concatenate 64 frames as one big image for VLM input. 
For 3D LLMs that take object-level point cloud instances as input, such as LEO, we set the number of instances to 40. We observe that increasing the instance number beyond 40 does not lead to a noticeable improvement in reasoning performance; therefore, to improve inference efficiency, we adopt this minimum number of instances.
For those models that encode the entire scene point cloud such as 3D-LLM, we uniformly downsample the scene point clouds to 5,000 points.

\subsection{Ablations of Different Modal Inputs}
\label{apdx:multimodal_ablation}

\begin{table}
\footnotesize
\centering
\caption{\textbf{Terms of different models and input modalities.}}
\begin{tabular}{@{}lll@{}}
\toprule
 \textbf{Model}  & \textbf{Input Modality} & \textbf{Term}  \\ \midrule
3D LLM/VLM/LLM  & Text-only & TI \\
3D LLM/VLM & Vision-only & VI  \\
3D LLM/VLM & Vision-text & VTI \\
3D LLM & Point cloud-only & PI  \\
3D LLM & Point cloud-vision & PVI\\
3D LLM & Point cloud-vision-text & PVTI \\
\bottomrule
\end{tabular}

\label{tab:terms}
\end{table}

To investigate the effect of multimodal inputs on 3D LLMs, we conduct controlled experiments by altering the input modalities of multimodal models. Since scenes can be represented in text, vision, or 3D point clouds, our key idea is to replace one modality with another (e.g., substituting point clouds with images or images with text). We design six modality combinations (Table~\ref{tab:terms}) for LLMs, VLMs, and 3D LLMs:
1) TI (Text-only): scene descriptions as input;
2) VI (Vision-only): multi-view images of the scene;
3) VTI (Vision-Text): multi-view images and scene descriptions;
4) PI (Point cloud-only): only point clouds as input, RGB values in point clouds and additional images are removed.;
5) PVI (Point cloud-Vision): Matches the original 3D LLM input, including point clouds with either RGB values or multi-view images;
6) PVTI (Point cloud-Vision-Text): Extends PVI by adding scene descriptions.
By default, PVI refers to the 3D LLM's original input.

\begin{figure*}[t]
  \centering
  \includegraphics[width=0.9\linewidth]{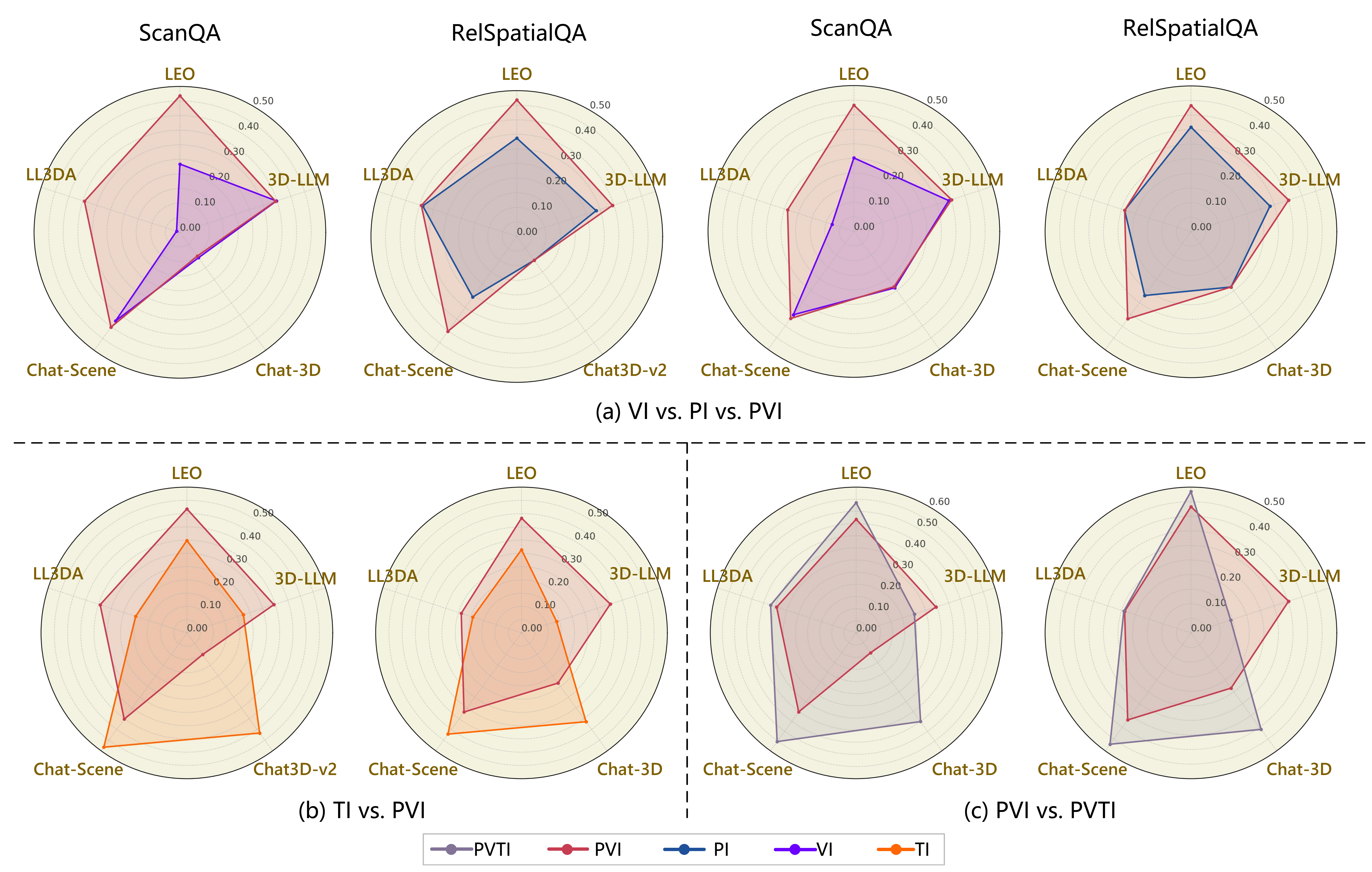}
  \caption{\textbf{Ablation studies with different modality inputs.}}
  \vspace{-0.4cm}
  \label{fig:multi_modal_ablation}
\end{figure*}

 We evaluate the performance of 3D LLMs on 3D QA tasks by feeding different modality combinations. For 3D LLMs that support visual input, such as LEO \cite{huang2023embodied} and 3D-LLM \cite{hong20233d}, we assess and compare their performance on 3D QA tasks using TQA-3D, VQA-3D, and PQA input. We assess 3D LLMs without visual input capabilities using only TQA-3D and PQA inputs. The results are depicted in Figure~\ref{fig:multi_modal_ablation}.

\textbf{The impact of point clouds and vision on the 3D LLMs depends on their training approach.
} To validate the effect of point clouds and visual signals on spatial reasoning, we evaluate models with VI and PI on ScanQA and RelSpatialQA datasets by masking the corresponding modality. To obtain PI, we mask the RGB values for models with RGB point clouds as PVI (LL3DA, LEO) and mask image features for models with point clouds and multi-view images as PVI (Chat-Scene, 3D-LLM). To obtain VI, we mask the coordinate values of the point cloud.
As depicted in Figure~\ref{fig:multi_modal_ablation} (a) where each vertex on the radar plot denotes a refined EM@1 of a 3D LLM, an interesting result is that after masking the coordinate values of point clouds, the performance of 3D-LLM, Chat-Scene, and Chat3D-v2 have been merely affected, while the performance of LL3DA and LEO have dropped significantly. LL3DA has nearly zero EM@1 with VI. When the point clouds are masked, the Chat-Scene and 3D-LLM suffer a performance drop. Though LEO also has a performance drop with PI, it is not as much as that with VI. This result indicates that 3D LLMs with multi-view images input are more sensitive to visual modality while those with RGB point clouds are more affected by the point cloud modality.

\textbf{Text-only input could lead to better performance than multi-modal input.} We conducted an ablation study evaluating 3D LLM with TI to validate the necessity of multi-modal input for spatial reasoning. As depicted in Figure~\ref{fig:multi_modal_ablation} (b), Chat-Scence and Chat-3D with VI outperform their PVI counterparts while LEO, LL3DA and 3D-LLMs lag behind their PVI performance. The results of Chat-Scene and Chat-3D demonstrate that the model solely relies on text input could even outperform multimodal input on 3D spatial reasoning tasks. In this study, point clouds fail to demonstrate their necessity in achieving the current SOTA performance on 3D spatial reasoning tasks. One possible explanation is that existing SOTA 3D LLMs with EM@1 $\textless$ 50 can correctly answer fewer than half of the questions in 3D QA, and these questions can be answered merely with visual perception and commonsense reasoning (e.g., identifying the color of a wooden cabinet).

\textbf{The redundancy in the text modality enhances the spatial reasoning performance.} Current 3D LLMs primarily use point clouds and images as inputs. While these modalities contain sufficient information for spatial reasoning tasks, according to~\cite{wang2025picture}, VLMs are more effective at leveraging textual descriptions for spatial reasoning. To investigate whether this conclusion extends to 3D LLMs, we evaluate their EM@1 scores using both PVI and PVTI inputs on the ScanQA and RelSpatialQA datasets. The results in Figure~\ref{fig:multi_modal_ablation} (c) clearly show that all 3D LLMs, except 3D-LLM, achieve higher EM@1 scores with PVTI than with PVI. This indicates that redundancy in the text modality enhances spatial reasoning performance, supporting the extension of this conclusion from VLMs to 3D LLMs.

\begin{table*}[t]
\centering
\caption{\textbf{EM@1 of different 3D LLMs with multimodal input combinations}.}
\resizebox{\linewidth}{!}{
\begin{tabular}{cccccc|ccccc}
\toprule
\multicolumn{6}{c|}{RelSpatialQA} & \multicolumn{5}{c}{ScanQA} \\
\midrule
Modality & 3D-LLM & LEO & LL3DA & ChatScene & Chat3D & 3D-LLM & LEO & LL3DA & ChatScene & Chat3D \\
\midrule
Mask All & 0.273 & 0.217 & 0.079 & 0.288 & 0.239 & 0.305 & 0.208 & 0.012 & 0.294 & 0.108 \\
TI & 0.141 & 0.315 & 0.193 & \textbf{0.473} & \textbf{0.414} & 0.224 & 0.349 & 0.203 & \textbf{0.555} & \underline{0.467} \\
VI & \underline{0.342} & 0.252 & 0.045 & 0.353 & 0.239 & \textbf{0.348} & 0.234 & 0.012 & 0.376 & 0.108 \\
VTI & 0.141 & 0.335 & 0.234 & 0.418 & \textbf{0.414} & 0.248 & 0.381 & 0.203 & 0.518 & \textbf{0.468} \\
PI & 0.285 & 0.359 & \textbf{0.243} & 0.270 & 0.234 & 0.286 & 0.337 & 0.339 & 0.257 & 0.104 \\
PTI & - & \underline{0.442} & - & - & 0.409 & - & - & \underline{0.369} & - & 0.451 \\
PVI & \textbf{0.352} & 0.435 & \underline{0.241} & 0.369 & 0.231 & \underline{0.345} & \underline{0.468} & 0.344 & 0.403 & 0.101 \\
PVTI & 0.143 & \textbf{0.483} & \underline{0.241} & \underline{0.472} & \underline{0.412} & 0.252 & \textbf{0.536} & \textbf{0.370} & \underline{0.553} & 0.456 \\
\bottomrule
\end{tabular}
\label{tab:total_abla}
}
\end{table*}

For complete results of models' EM@1 with different combinations of input modalities, please refer to Table~\ref{tab:total_abla}.

\subsection{Data Annotation}
We employed five student annotators for data refinement and compensated them at a rate of ¥0.5 per instance. Prior to annotation, we provided three refinement guidelines: (1) filter out irreversible spatial triplets and their corresponding QA pairs, (2) correct inaccurate triplets, and (3) revise grammatical errors in QA cases. After five days of annotation, a total of 5,523 high-quality QA samples were obtained.

\subsection{LLM Usage}
LLMs were used solely to assist with manuscript preparation. Their contributions were limited to language editing tasks, such as improving clarity, readability, grammar, and overall writing quality. All scientific aspects of the work, including the research ideas, methodology, experiments, analyses, and conclusions, were conceived and conducted exclusively by the authors. The LLMs did not contribute to the generation of research content or the interpretation of results.

We take full responsibility for all content presented in this manuscript, including portions that were edited or refined using LLM assistance.
\end{document}